\documentclass[12pt,a4paper]{article}
\usepackage{placeins}
\usepackage[margin=1in]{geometry}
\usepackage{setspace}
\usepackage{float}
\usepackage{titlesec}
\usepackage{graphicx}
\usepackage{pdflscape}
\usepackage{amsthm}
\usepackage{times}
\usepackage{amsmath,amssymb}
\usepackage{booktabs}
\usepackage{subcaption}
\usepackage{multirow}
\usepackage[hidelinks]{hyperref}

\newtheorem{conjecture}{Conjecture}
\titleformat{\section}
  {\bfseries\large}
  {\thesection.}
  {0.5em}
  {}

\begin{document}

\begin{center}

{\LARGE\bfseries
Functional Equivalence and Geometric Diversity in Neural Network Approximations: An Empirical Characterization}

\vspace{1.2em}

{\large
S. A. Anuragine$^{1}$,
Prem Jagadeesan$^{1,*}$
}

\vspace{0.8em}

{\normalsize
$^{1}$Amrita School of Artificial Intelligence\\
Amrita Vishwa Vidyapeetham, Coimbatore, India
}

\vspace{0.8em}

{\small
E-mail: cb.ai.u4aim24005@cb.amrita.students.edu \\
*Corresponding author: j\_prem@cb.amrita.edu
}

\end{center}

\vspace{1em}

\section*{Abstract}
The Universal Approximation Theorem states that a neural network with a single hidden layer is sufficient to approximate any continuous univariate function on a compact domain to arbitrary error. However, the uniqueness of such neural network representations is not guaranteed, raising questions about practical identifiability. In this work, we address this concern by analyzing functional equivalence and geometric diversity of neural network approximations to a few elementary mathematical functions. The analysis includes an extensive study of single-layer neural networks and multilayer perceptrons under noisy and noise-free conditions. Beyond just network capacity, we study the geometric properties through the lens of sloppiness, characterized by the eigen spectrum of the Hessian of the cost function and the effective rank to quantify the dimensionality of parameter space. The study reveals large equivalence classes of functionally indistinguishable yet geometrically diverse networks that consistently exhibit low effective rank and structural redundancy. Finally, a model select criterion is proposed for identifying optimal models based on parsimony, ease of estimation, and inference efficiency.

\subsubsection*{Keywords}
Neural Network Approximations, Functional equivalence, Geometric diversity, Effective rank

\section{Introduction}

In data-driven modeling, neural networks are central to many applications. They are widely used across image analysis, natural language processing and generative artificial intelligence (\cite{article}, \cite{Goodfellow-et-al-2016}). However, these models are found to be overparameterized and redundant, which of results in a computational bottleneck for many downstream tasks. 
There may exist multiple network architectures of varying capacities that give rise to nearly identical approximations. However,  
in practice, we do not have a systematic guiding principle for selecting a parsimonious network structure. This often leads to increasing the number of layers, neurons, or trainable parameters until the desired approximation accuracy is achieved.

This lack of a guiding principle for network selection leads to over-parameterization and redundancy. When a model contains more parameters than are effectively required, several distinct sub-network configurations may produce nearly identical fit. Therefore, minimizing  the approximation error alone may not always optimize the network capacity. This has resulted in multiple deep learning compression and pruning techniques as an attempt to arrive at a smaller network that suits the given application. These methods may be applied before, during, or after training, and they often rely on criteria such as weight magnitude, gradient information, or Hessian-based sensitivity (\cite{cheng2024surveydeepneuralnetwork} \cite{19ashkboos2024slicegptcompresslargelanguage} \cite{20ma2023llmprunerstructuralpruninglarge}) . Knowledge distillation is another model-compression approach in which a smaller student network is trained to imitate the behaviour of a larger teacher network (\cite{hinton2015distillingknowledgeneuralnetwork}) . Similarly, LoRA reduces the number of trainable parameters by freezing the original model weights and learning only small low-rank update matrices during fine-tuning (\cite{hu2021loralowrankadaptationlarge} \cite{21denton2014exploitinglinearstructureconvolutional} \cite{8478366}). However, these compression techniques are usually conditioned on an already selected or trained model, and therefore may not fully reveal the broader class of alternative architectures that could approximate the same target function equally well . 

Kolmogorov-Arnold Networks (KANs) have recently been proposed as an alternative architecture inspired by the Kolmogorov-Arnold representation theorem, and have shown promise in achieving compact representations with fewer parameters than traditional neural networks (\cite{liu2025kankolmogorovarnoldnetworks},\cite{liu2025kankolmogorovarnoldnetworks}). However, their reliance on spline-based nonlinear transformations can introduce high computational cost and scalability limitations for larger models (\cite{tran2024exploringlimitationskolmogorovarnoldnetworks}). Similarly, the Lottery Ticket Hypothesis suggests that dense randomly initialized networks contain smaller subnetworks, called winning tickets, which can be trained in isolation to achieve performance comparable to the original network (\cite{frankle2019lotterytickethypothesisfinding}). While all these works point towards the existence of useful subnetworks, it still begins from an identified dense model and focuses on identifying sparse subnetworks within it. These continued attempts for finding a smaller network raises an important question: for a given target function, do multiple neural networks exist that give nearly identical fit and behave functionally similar.

In this work, we take a different perspective. Instead of beginning with one optimal network and compressing it, we study the existence of multiple functionally equivalent networks by defining an equivalence class (\cite{rudin1987real}) and analyze how their parameter-space geometries differ. 
The primary reason for analyzing the geometry of the parameter space is that it affects the ease of estimation and therefore, it dictates the resulting architecture. 
The geometric analysis is carried out through the lens of sloppiness. A model is considered sloppy when there exist large regions in the parameter space over which model predictions remain nearly identical. Such behavior can be studied using the eigenvalue spectrum of the Hessian of the loss function, where large eigenvalues correspond to stiff directions and small eigenvalues correspond to sloppy or weakly constrained directions (\cite{sagun2017eigenvalueshessiandeeplearning}, \cite{10.1371/journal.pcbi.0030189}, \cite{10.1371/journal.pone.0282609}). In addition, the effective rank of the Hessian is used to estimate how many parameter-space directions contribute meaningfully to the local parameter space geometry and can be considered as an estimate of sloppiness (\cite{1585},\cite{https://doi.org/10.1002/cem.874},\cite{7098875}).

Recent studies on neural-network Hessians show that the Hessian is not an arbitrary high-dimensional matrix, but often contains a structured low-rank dominant subspace (\cite{wu2022dissectinghessianunderstandingcommon}). Another line of work explains the spectral behavior of Hessian of neural networks by relating it to the structure present in the data and learned representations. It shows that matrices such as the Hessian and Fisher information matrix often contain a few dominant eigenvalue directions along with a large bulk of smaller directions (\cite{papyan2020tracesclasscrossclassstructurepervade}). 
While the existing literature studies mainly analyze Hessian structure for selected trained networks/tasks or specific model instances, this work examines the phenomenon across a complete model space under the chosen architectural constraints.

While modern architectures such as Transformers (\cite{vaswani2023attentionneed}) and KANs operate on high-dimensional parameter space, the scope of this foundational work is deliberately focused on the approximation of univariate elementary mathematical functions using single-layer and multi-layer feedforward networks. Studying these univariate approximations serves as an essential first step toward decoding the fundamental nature of neural network approximations. Across these experiments, several networks are observed to achieve nearly identical approximation performance forming an equivalence class. The multilayer perceptron analysis further shows that functional equivalence is not restricted to a few models, but occupies a significant portion of the model space under the chosen architectural constraints. Even when the networks differ in model capacity, their effective rank remains much smaller than the total number of trainable parameters, indicating that only a limited portion of the parameter space contributes strongly to the learned representation, this suggests that the interaction of data and the model structure induces a low-dimensional geometry in the parameter space and the network always remains over-parameterized. Similar behavior is also observed under noisy conditions, suggesting that the low-dimensional geometric structure is not removed by perturbations in the data.

Thus, neural network approximation is not only a question of whether a function can be learned, but also of how many distinct networks can learn it and how much of their parameter space is actually relevant to the representation. The study of geometry of parameter space offers rich insights into the neural network approximations and resulting equivalence class is therefore functionally similar, geometrically diverse, and constrained by a low-dimensional effective geometry.

\section{Experimental Design}

The proposed methodology consists of two experiments. The first investigates shallow neural networks by varying the number of hidden neurons and activation functions to identify functionally equivalent approximations. The second extends the study to multilayer perceptrons (MLPs) with varying depths and widths to examine functional equivalence across a larger architecture space. The detailed network configurations and data generation procedure are provided in Supplementary $S2$.

For every network, the parameters are estimated and the networks are ranked according to approximation performance. The top-$K$ networks are selected to form an equivalence class. To evaluate robustness, the complete MLP experiment is repeated under both noise-free and noisy conditions.

\section{Results}
Let $f_{\theta_1}$ and $f_{\theta_2}$ denote two neural network approximations belonging to a normed function space $\mathcal{F}$. The networks are said to be functionally equivalent with tolerance $\varepsilon$ if

\begin{equation}
\label{eq:1}
\|f_{\theta_1} - f_{\theta_2}\|_{\mathcal{F}} < \varepsilon .
\end{equation}

In this work, functional equivalence is assessed using approximation metrics such as $R^2$ and NRMSE. Inspired by \cite{rudin1987real},
we define the equivalence class of a network $f_{\theta}$ by

\begin{equation}
\label{eq:2}
[f_{\theta}]_{\varepsilon}
=
\left\{
f_{\theta'} :
\|f_{\theta'} - f_{\theta}\| < \varepsilon
\right\}.
\end{equation}

So all networks $f_{\theta}$ that satisfy the approximation criterion defined by \eqref{eq:1} are considered to be functionally non-unique approximations. Having defined \eqref{eq:2}, the core idea of this work is to study the geometric characteristics of the functionally non-unique networks using the concept of sloppiness.

\subsection{Analysis of Single Layer Neural Networks}
This section presents the results of the empirical study on shallow neural networks (single-layer). Figure~\ref{fig:combined}(a) depicts functional non-uniqueness and geometric diversity of a class of 25 neural networks fit to a sinusoidal function with hyperbolic tangent activation. The networks are ranked on the basis of $R^2$ measure. It is evident from Figure ~\ref{fig:combined}(a) that top 25 neural networks have numerically indistinguishable $R^2$ and $\mathrm{NRMSE}$ less than an arbitrarily small $ \varepsilon$, making them practically unidentifiable network structures. However, the singular value ratio varies substantially across these networks leading to significant geometric diversity. Despite network capacities ranging from 16 to 151 parameters, the effective rank remains between 0.5591 and 1.8056. This suggests that only a small subset of parameter-space directions dominate the local geometry. This makes effective rank a noteworthy geometric property of the parameter space.

It is observed that the geometric properties of the parameter space are functions of data, network structure and activation function.
Table~\ref{tab:network_comparison} summarizes the network structures obtained for different target function--activation combinations along with their effective rank, singular value ratio, and Hessian eigenvalue spectrum. Although these networks achieve nearly identical approximation performance, they exhibit noticeable differences in their geometric properties, demonstrating that functional equivalence does not imply geometric similarity. In particular, the consistently low effective rank indicates that only a small subset of the parameter space contributes significantly to the learned representation, while the remaining directions are largely redundant. Furthermore, the effective rank varies across activation functions, suggesting that the choice of activation influences the geometry of the parameter space.

Figure~\ref{fig:combined}(a) further illustrates that the equivalent networks produce nearly indistinguishable function approximations despite substantial differences in their parameter-space geometry. These observations provide evidence for the existence of functionally equivalent yet geometrically diverse neural network structures. Detailed results for each target function and activation function combination, including approximation, geometric diversity, and Hessian eigenspectrum analyses, are available in the Supplementary Material (Figures S1--S16).  

\subsection{Analysis of Multilayer Perceptron}

In this section, we extend the analysis to multilayer perceptrons (MLP) to verify that the observations are not an artifact of single layer networks. The analysis is done for a set of 3905 networks whose number of layers and neurons in each layer varies from 1 to 5.

\begin{figure*}[!t]
\centering


\begin{subfigure}[t]{0.47\textwidth}
    \centering
    \includegraphics[
        width=\linewidth,
        trim={0cm 0cm 0cm 0cm},
        clip
    ]{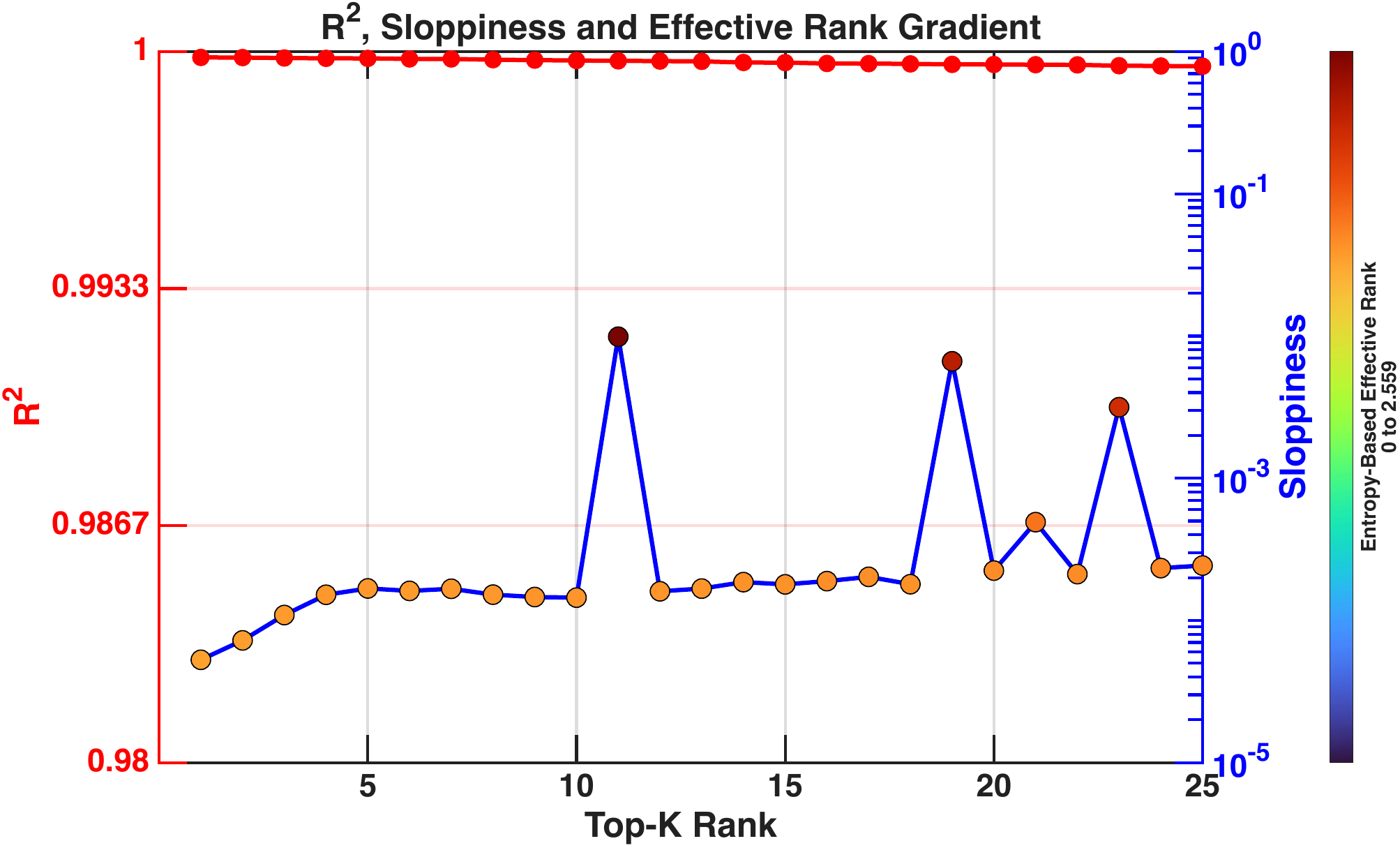}
    \caption{}
    \label{fig:1}
\end{subfigure}
\hfill
\begin{subfigure}[t]{0.47\textwidth}
    \centering
    \includegraphics[
        width=\linewidth,
        trim={0cm 19.5cm 0cm 18cm},
        clip
    ]{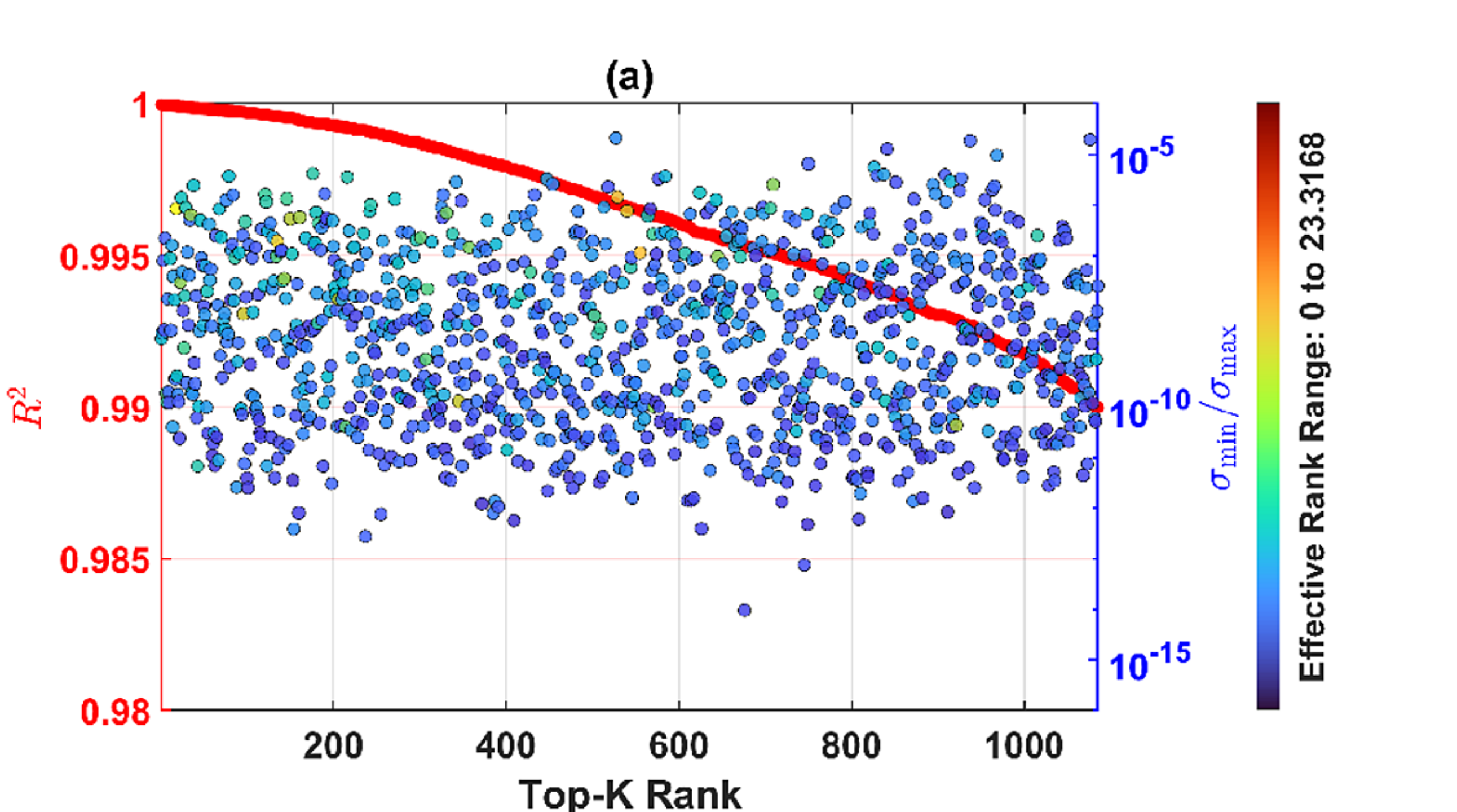}
    \caption{}
    \label{fig:2}
\end{subfigure}

\vspace{0.4em}


\begin{subfigure}[t]{0.47\textwidth}
    \centering
    \includegraphics[
        width=\linewidth,
        trim={0cm 19.5cm 0cm 18cm},
        clip
    ]{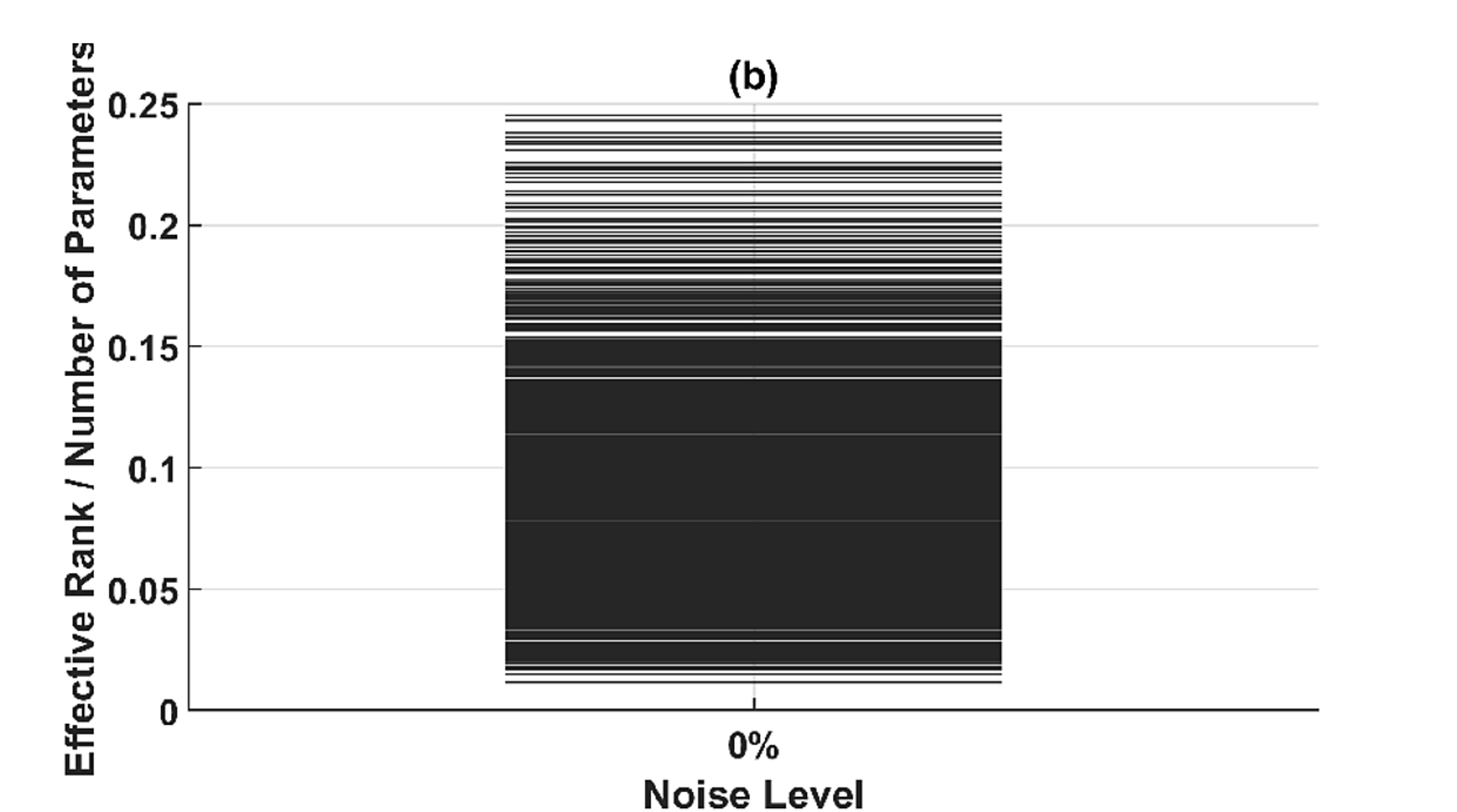}
    \caption{}
    \label{fig:3}
\end{subfigure}
\hfill
\begin{subfigure}[t]{0.47\textwidth}
    \centering
    \includegraphics[
        width=\linewidth,
        trim={0cm 19.5cm 0cm 18cm},
        clip
    ]{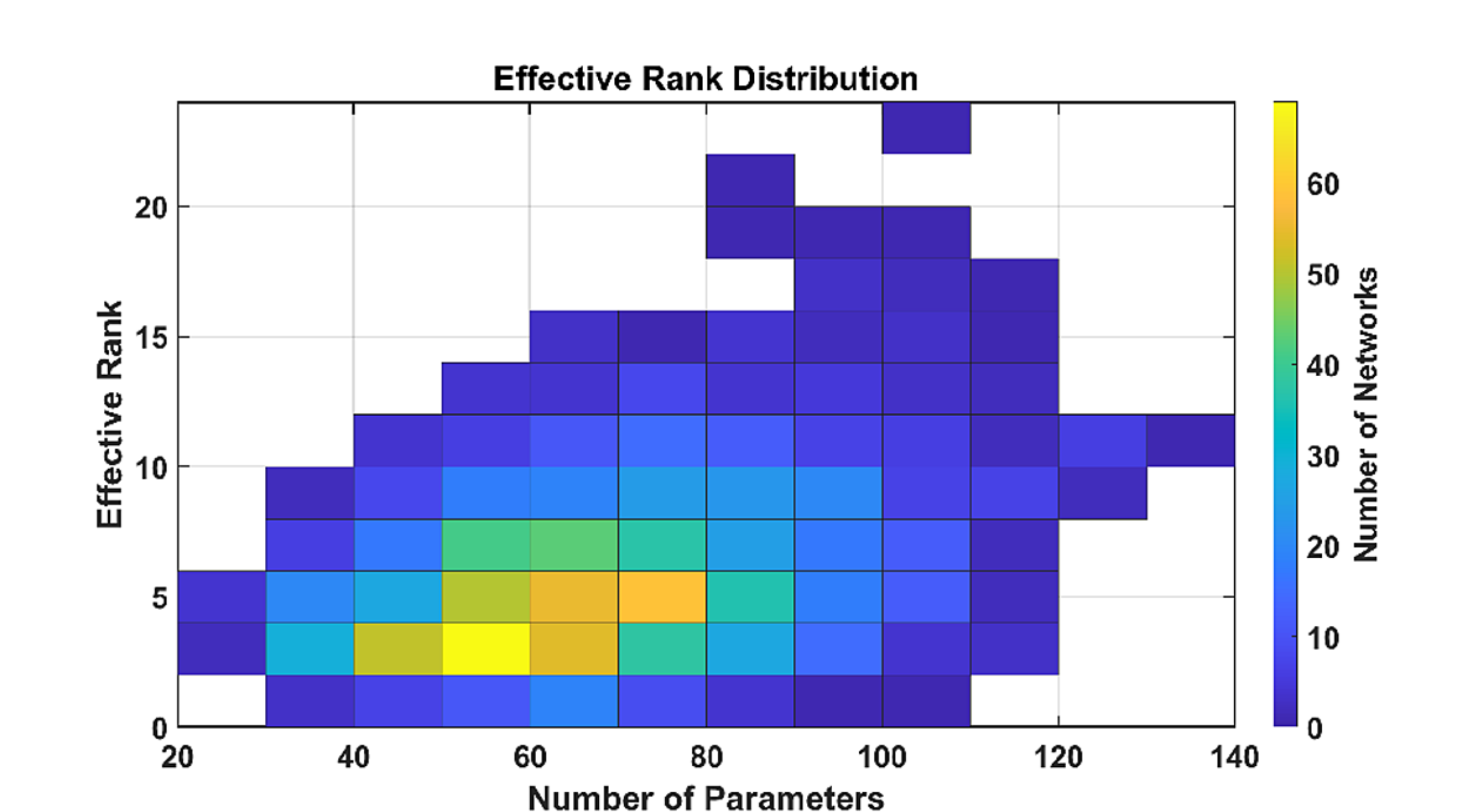}
    \caption{}
    \label{fig:4}
\end{subfigure}

\vspace{0.4em}


\begin{subfigure}[t]{0.47\textwidth}
    \centering
    \includegraphics[
        width=\linewidth,
        trim={0cm 19.5cm 0cm 18cm},
        clip
    ]{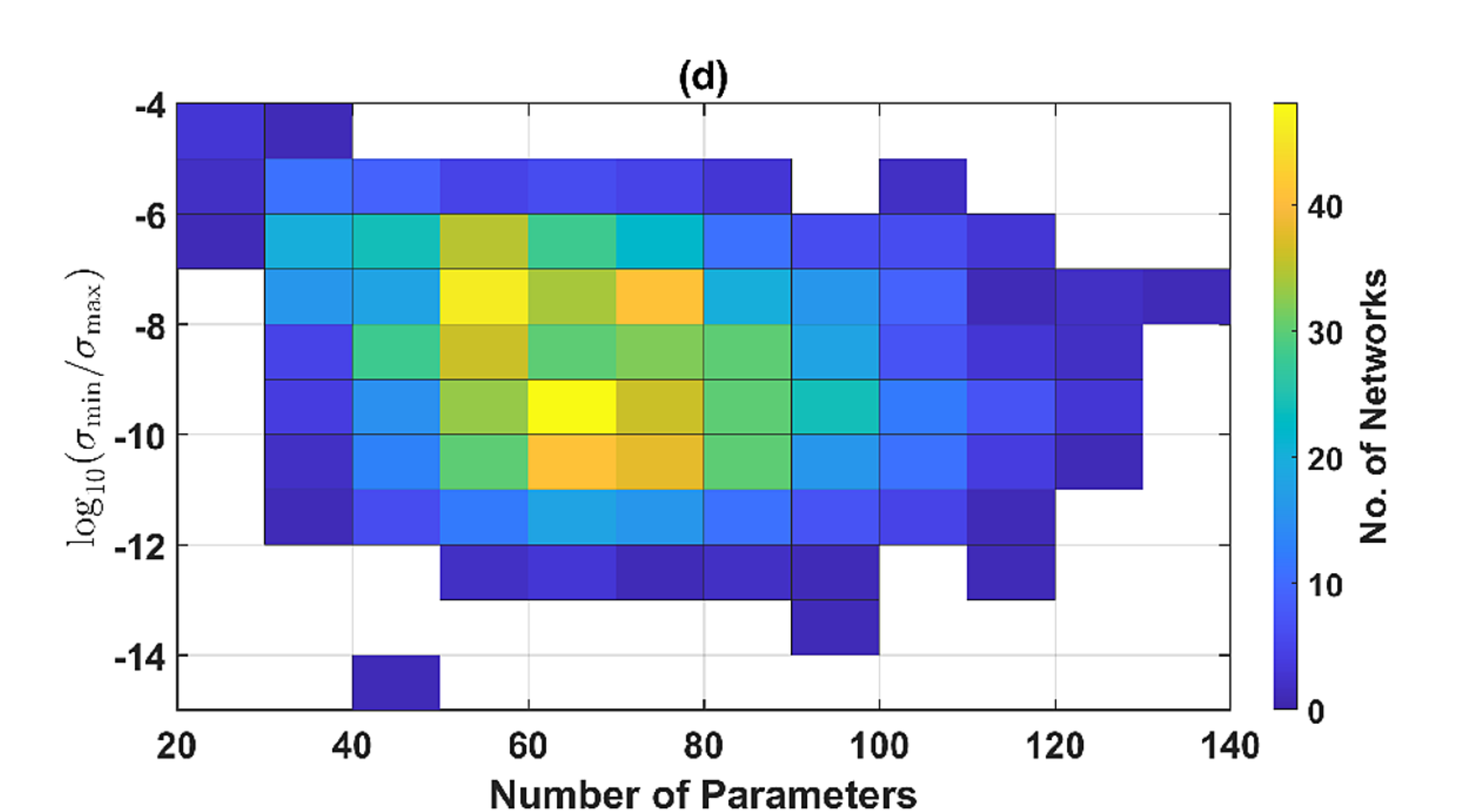}
    \caption{}
    \label{fig:5}
\end{subfigure}
\hfill
\begin{subfigure}[t]{0.47\textwidth}
    \centering
    \includegraphics[
        width=\linewidth,
        trim={0cm 19.5cm 0cm 18cm},
        clip
    ]{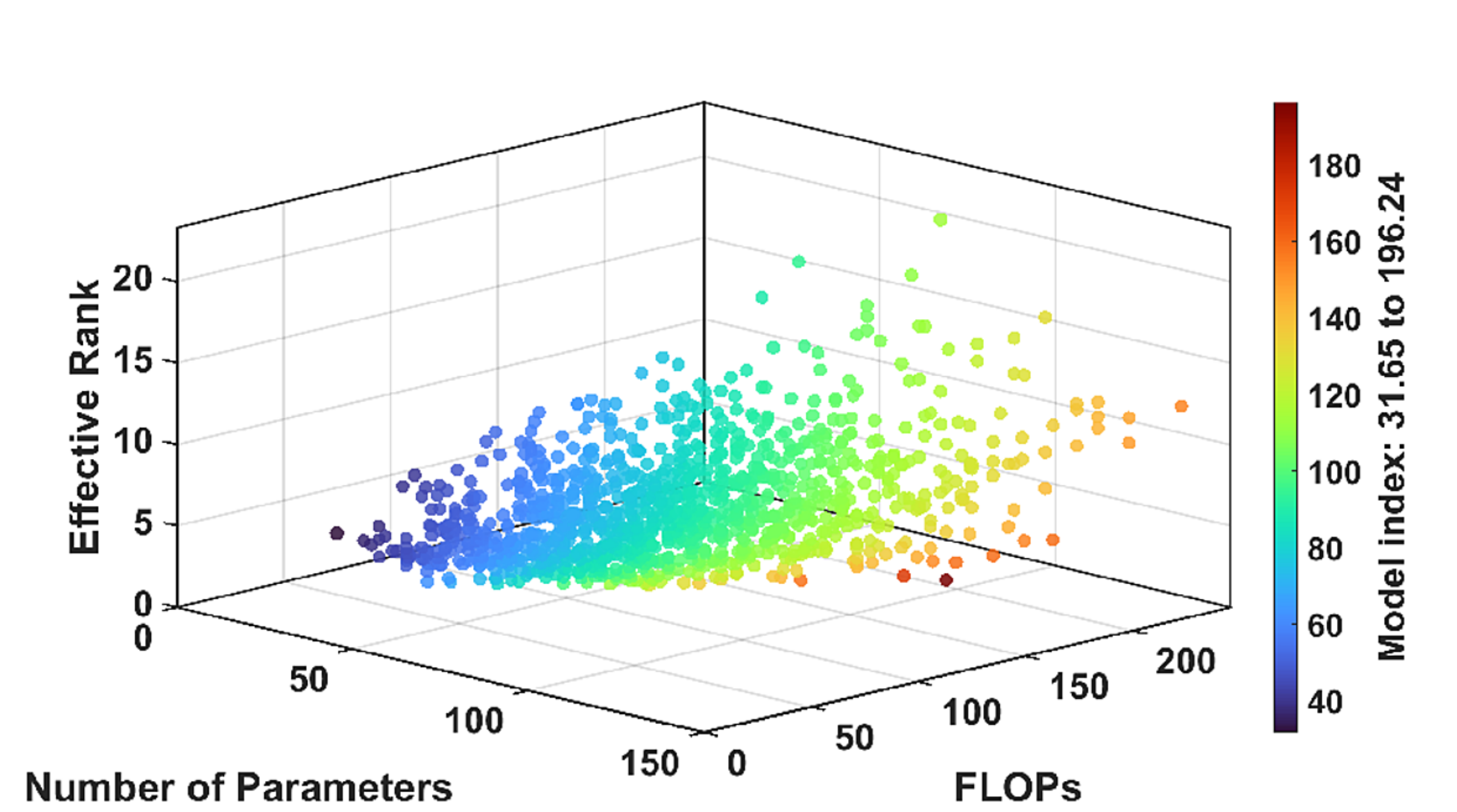}
    \caption{}
    \label{fig:6}
\end{subfigure}

\caption{Summary of functional equivalence, geometric characterization, and model selection in neural network approximations. The figure illustrates the existence of functionally equivalent architectures, their geometric diversity in parameter space, and the resulting implications for selecting computationally efficient models.}

\label{fig:combined}

\end{figure*}

For the noise-free scenario, the equivalence class consists of 1085 networks with \(R^2 \geq 0.99\), indicating that a substantial fraction of the architecture space belongs to the equivalence class. Figure~\ref{fig:combined} summarizes the geometric characteristics of these networks. Figure~\ref{fig:combined}(b) shows that the equivalent networks achieve numerically indistinguishable \(R^2\) values, demonstrating functional non-uniqueness. Figure~\ref{fig:combined}(c) presents the ratio of effective rank to the total number of parameters, revealing an upper bound of approximately 0.25 across all architectures. This indicates that only a small fraction of the parameter space contributes significantly to the learned representation, while the remaining directions are redundant. Figure~\ref{fig:combined}(d) shows that a large proportion of the equivalent networks are concentrated within a relatively narrow range of parameter counts and effective ranks. Likewise, Figure~\ref{fig:combined}(e) shows that most equivalent networks exhibit singular value ratios between \(10^{-11}\) and \(10^{-6}\), indicating that they are highly sloppy. Overall, these observations demonstrate that neural network approximations admit a large equivalence class of functionally indistinguishable architectures whose parameter-space geometries differ substantially, while their effective dimensionality remains constrained to a significantly lower-dimensional subspace.
To examine the effect of noise on functional equivalence and geometric diversity, the same set of 3905 MLP architectures was evaluated after adding Gaussian noise to the target data. Although the cardinality of the equivalence class decreases considerably, the overall geometric behaviour remains unchanged. In particular, the ratio of effective rank to parameter count continues to remain below approximately 0.25, and the equivalent networks exhibit similarly small singular value ratios, indicating that the low-dimensional and sloppy nature of the parameter space persists even under noisy conditions. The complete analysis for the noisy case is provided in Supplementary S4.1.

Overall, approximating noisy data to neural networks significantly reduces the cardinality of the equivalence class while significant diversity exists in the parameter space geometry.

\FloatBarrier
\begin{table}[H]
\centering
\scriptsize
\setlength{\tabcolsep}{3pt}

\caption{Comparison of Top-K Networks for Different Functions and Activation Functions}
\vspace{0cm}

\resizebox{\textwidth}{!}{

\begin{tabular}{|c|c|c|c|c|c|c|c|c|c|c|c|c|c|}
\hline

{\textbf{Function}} &
{\textbf{Activation}} &

\multicolumn{2}{c|}{\textbf{No. of Neurons}} &
\multicolumn{2}{c|}{\textbf{No. of Parameters}} &
\multicolumn{2}{c|}{\textbf{Effective Rank}} &
\multicolumn{2}{c|}{\textbf{Singular value ratio}} &
\multicolumn{2}{c|}{\textbf{$R^2$}} &
\multicolumn{2}{c|}{\textbf{NRMSE}} \\

\cline{3-14}

& &
\textbf{Min} & \textbf{Max} &
\textbf{Min} & \textbf{Max} &
\textbf{Min} & \textbf{Max} &
\textbf{Min} & \textbf{Max} &
\textbf{Min} & \textbf{Max} &
\textbf{Min} & \textbf{Max} \\

\hline

Sin
& Tanh
& 5 & 50
& 16 & 151
& 1.8056 & 2.5591
& $2\times10^{-5}$ & $1\times10^{-12}$
& 0.99959 & 0.99984
& 0.004 & 0.007 \\
\cline{2-14}

& Sigmoid
& 13 & 50
& 40 & 151
& 2.403 & 2.5358
& $8\times10^{-9}$ & $2\times10^{-12}$ 
& 0.99937 & 0.99984
& 0.004 & 0.008 \\
\cline{2-14}

& ReLU
& 30 & 75
& 91 & 226
& 5.9559 & 95.045
& $0$ & $5\times10^{-10}$ 
& 0.99089 & 0.99981
& 0.004 & 0.033 \\
\hline

Cos
& Tanh
& 5 & 50
& 16 & 151
& 1.8081 & 2.4962
& $6\times10^{-6}$ & $3\times10^{-13}$ 
& 0.9995 & 0.99984
& 0.004 & 0.007 \\
\cline{2-14}

& Sigmoid
& 26 & 50
& 79 & 151
& 2.4056 & 2.5369
& $2\times10^{-10}$ & $4\times10^{-12}$ 
& 0.99932 & 0.99983
& 0.004 & 0.009 \\
\cline{2-14}

& ReLU
& 20 & 74
& 61 & 223
& 10.756 & 105.72
& $0$ & $2\times10^{-19}$ 
& 0.99453 & 0.99905
& 0.01 & 0.026 \\
\hline

Exp
& Tanh
& 2 & 50
& 7 & 151
& 2.516 & 5.4948
& $4\times10^{-4}$ & $7\times10^{-14}$ 
& 0.99994 & 0.99999
& 0.0007 & 0.001 \\
\cline{2-14}

& Sigmoid
& 2 & 50
& 7 & 151
& 2.5723 & 3.5663
& $2\times10^{-4}$ & $1\times10^{-12}$ 
& 0.99992 & 0.99984
& 0.0009 & 0.002 \\
\cline{2-14}

& ReLU
& 34 & 75
& 103 & 226
& 23.321 & 75.752
& $0$ & $1.4\times10^{-24}$ 
& 0.99999 & 1
& 0.0004 & 0.0008 \\
\hline

Quadratic
& Tanh
& 2 & 50
& 7 & 151
& 2.8333 & 3.5274
& $1\times10^{-3}$ & $9\times10^{-13}$ 
& 0.99984 & 0.99995
& 0.002 & 0.003 \\
\cline{2-14}

& Sigmoid
& 2 & 50
& 7 & 151
& 2.3837 & 3.2462
& $6\times10^{-3}$ & $4\times10^{-12}$ 
& 0.99981 & 0.99994
& 0.002 & 0.004 \\
\cline{2-14}

& ReLU
& 34 & 75
& 103 & 226
& 11.084 & 26.243
& $0$ & $6\times10^{-25}$ 
& 0.99997 & 0.99999
& 0.0008 & 0.001 \\
\hline

Cubic
& Tanh
& 3 & 50
& 10 & 151
& 3.0306 & 3.7035
& $4\times10^{-5}$ & $2\times10^{-18}$ 
& 0.99993 & 1
& 0.0003 & 0.001 \\
\cline{2-14}

& Sigmoid
& 3 & 50
& 10 & 151
& 2.2249 & 3.186
& $8\times10^{-6}$ & $7\times10^{-21}$ 
& 0.99991 & 0.99999
& 0.0007 & 0.001 \\
\cline{2-14}

& ReLU
& 36 & 75
& 109 & 226
& 19.285 & 70.213
& $0$ & $5\times10^{-24}$ 
& 0.99997 & 0.99999
& 0.0005 & 0.001 \\
\hline

\end{tabular}

}

\label{tab:network_comparison}

\end{table}
\newpage
\section{Prem-Anu Selection Criteria}
In this section, we attempt to answer one of the most important questions in data-driven modeling. We show in the previous section that neural network approximations form an equivalent class with varying geometric properties across architectures. This leads to the crucial question: which of these models are good? To answer this, we introduce a model index ($\mathcal{M_I}$) defined by (\ref{conjecture}) which is a function of model capacity, effective rank and FLOPs. 

\begin{equation}
    \mathcal{M_I}= Np \big(1+\frac{1}{r_{eff}}\big) + \log F 
    \label{conjecture}
\end{equation}
where $Np$ is number of model parameters, $r_{eff}$ is the effective rank of the Hessian of the cost function and $F$ is the FLOPs. The proposed model index is a heuristic motivated by the observations reported in this study and is intended as a first step in the right direction of  principled model selection within an equivalence class

\begin{conjecture}
Let $[f_{\theta}]_{\varepsilon}$ be an equivalence class of all practically unidentifiable neural network structures defined by \eqref{eq:2} for a given target function $f$. The optimal model in terms of ease of estimation, inference energy and parsimony is the one that minimizes the model index $\mathcal{M_I}$.

\end{conjecture}
The core idea behind calculating $\mathcal{M_I}$ is that a model that minimises this index inherently minimises the number of parameters (parsimony), minimises the difference between the number of parameters and the effective rank (sloppiness), which in turn improves ease of estimation and finally FLOPs, inference time. Thus, $\mathcal{M_I}$ can serve as a useful index to find an optimal model among a set of functionally equivalent models. We show numerical evidence for the proposed conjecture in Figure \ref{fig:combined}(f). The proposed criterion is illustrated by analyzing equivalent models under noise-free and noisy conditions (supplementary S4.2), where lower model index corresponds to architectures with fewer parameters, lower effective rank, and reduced FLOPs.    

\section{Discussion}
In this study, we address the question of functional non-uniqueness and geometric diversity in neural network approximations. We show that functionally equivalent network structures exhibit significant diversity in parameter-space geometry, characterized by the Hessian eigenvalue spectrum, effective rank, and singular value ratio, irrespective of network capacity. These observations are invariant to random parameter initializations. The MLP analysis further reinforces these findings. Although the addition of noise reduces the size of the equivalence class, its geometric characteristics remain largely unchanged. Across all equivalent networks, only a small fraction of the parameter space contributes meaningfully to the learned representation, while the remaining directions are redundant, as reflected by the consistently low effective rank. This suggests that the low-dimensional geometry is governed primarily by the interaction between the target function and the network architecture, and that increasing the number of parameters expands the parameter space without increasing its effective dimensionality.

These findings have implications beyond functional equivalence. The existence of large equivalence classes indicates that multiple networks can approximate the same target function. Interestingly, even the smallest network within the equivalence class contains redundant parameters, motivating the search for parsimonious architectures. Since only a low-dimensional subset of parameters governs the learned representation, parameter estimation can be viewed through these effective directions rather than the full parameter space. Thus, the challenge is not merely to approximate a target function accurately, but to identify the simplest network satisfying important criteria such as parsimony, ease of estimation, and inference time. The proposed conjecture provides a first step towards selecting such an optimal model from the equivalence class. An important direction for future research is to determine whether an even smaller network can be derived from the smallest equivalent model and to understand the origin of this redundancy through the geometry of the parameter space. 

\section*{Author Contributions}

\noindent\textbf{Conceptualization:}  Prem Jagadeesan.\\
\textbf{Formal analysis:} Anuragine S A, Prem Jagadeesan.\\
\textbf{Methodology:} Prem Jagadeesan.\\
\textbf{Supervision:} Prem Jagadeesan.\\
\textbf{Visualization:} Anuragine S A, Prem Jagadeesan.\\
\textbf{Writing -- original draft:} Anuragine S A, Prem Jagadeesan.\\
\textbf{Writing -- review \& editing:} Anuragine S A, Prem Jagadeesan.\\.

\nocite{*}
\bibliographystyle{unsrt}
\bibliography{Ref}

\renewcommand{\thesection}{S\arabic{section}}
\renewcommand{\thesubsection}{S\arabic{section}.\arabic{subsection}}

\renewcommand{\thefigure}{S\arabic{figure}}
\renewcommand{\thetable}{S\arabic{table}}
\renewcommand{\theequation}{S\arabic{equation}}
\newpage
\begin{center}

{\LARGE\bfseries Supplementary Material}

\vspace{0.4cm}

for

\vspace{0.4cm}

{\Large\bfseries
Functional Equivalence and Geometric Diversity in Neural Network Approximations: An Empirical Characterization
}

\vspace{0.7cm}

S. A. Anuragine$^{1}$,
Prem Jagadeesan$^{1}$

\vspace{0.3cm}

{\normalsize$^{1}$Amrita School of Artificial Intelligence\\Amrita Vishwa Vidyapeetham, Coimbatore, India}

\vspace{0.8em}

\end{center}

\vspace{1cm}

\tableofcontents

\newpage

\section{Definitions and Metrics}

This summarizes the principal metrics and concepts employed throughout the paper for analyzing and evaluating neural network approximations, geometric diversity, and model complexity.

\subsection{Sloppiness}

A neural network is said to exhibit \emph{sloppiness} when the eigenvalues of its Hessian span several orders of magnitude. In such models, the loss function is highly sensitive along a small number of stiff directions while remaining comparatively insensitive along many sloppy directions.

Let

\begin{equation}
H=\nabla^{2}\mathcal{L}(\theta),
\end{equation}

where $\mathcal{L}(\theta)$ denotes the training loss and $\theta$ represents the vector of trainable parameters.

If

\[
\lambda_1\ge\lambda_2\ge\cdots\ge\lambda_n,
\]

are the eigenvalues of $H$, then a broad spread of these eigenvalues is indicative of sloppy parameter-space geometry.

\subsection{Hessian Computation}

The local geometry of the parameter space is characterized by the Hessian matrix of the least-squares loss evaluated at the optimal parameter vector $\theta^{*}$. Itis approximated numerically using central finite differences. The $(i,j)$-th element of the Hessian is computed as

\begin{equation}
H_{ij}(\theta^{*})
=
\frac{
L(\theta^{*}+\epsilon e_i+\epsilon e_j)
-
L(\theta^{*}+\epsilon e_i-\epsilon e_j)
-
L(\theta^{*}-\epsilon e_i+\epsilon e_j)
+
L(\theta^{*}-\epsilon e_i-\epsilon e_j)
}
{4\epsilon^{2}},
\label{eq:numerical_hessian}
\end{equation}

where $L(\theta)$ denotes the least-squares loss function, $\epsilon$ is a small perturbation constant, and $e_i$ and $e_j$ are the standard basis vectors corresponding to the $i$-th and $j$-th parameters, respectively. To eliminate numerical asymmetry arising from finite-difference approximation, the Hessian is symmetrized as

\begin{equation}
H \leftarrow \frac{H+H^{T}}{2}.
\end{equation}

The eigenvalue spectrum of the resulting Hessian is subsequently used to compute the effective rank and singular value ratio, which characterize the local geometry and sloppiness of the parameter space.

\subsection{Effective Rank}

The entropy-based effective rank measures the intrinsic dimensionality of a matrix spectrum by interpreting the normalized singular values (or eigenvalues) as a probability distribution. Unlike the algebraic rank, the effective rank provides a continuous estimate of the number of dominant spectral directions contributing to the model.

It is defined as

\begin{equation}
\operatorname{erank}(A)
=
\exp\left(
-\sum_i p_i\log p_i
\right),
\end{equation}

where

\begin{equation}
p_i
=
\frac{\sigma_i}
{\sum_j \sigma_j},
\end{equation}

and $\sigma_i$ denotes the singular values (or absolute eigenvalues) of $A$.

\subsection{Normalized Root Mean Squared Error (NRMSE)}

The normalized root mean squared error (NRMSE) measures approximation accuracy while accounting for the scale of the target data. Lower values correspond to superior approximations.

\begin{equation}
\mathrm{NRMSE}
=
\frac{
\sqrt{\dfrac1N
\sum_{i=1}^{N}
(y_i-\hat y_i)^2}
}
{y_{\max}-y_{\min}}.
\end{equation}

\subsection{\texorpdfstring{$R^2$}{R2} Score}

The coefficient of determination quantifies the proportion of variance explained by the neural network approximation.

\begin{equation}
R^2
=
1-
\frac
{\sum_{i=1}^{N}(y_i-\hat y_i)^2}
{\sum_{i=1}^{N}(y_i-\bar y)^2},
\end{equation}

where $\bar y$ denotes the mean of the target values.

Values closer to one indicate improved approximation quality.

\FloatBarrier

\section{Experimental Details}

This provides the complete experimental settings required to reproduce all numerical results presented in the main manuscript.

\subsection{Data Generation}

Datasets are generated from elementary univariate mathematical functions, including trigonometric, exponential, and polynomial functions. Sampling intervals are selected to sufficiently capture the behaviour of each target function while maintaining uniform sampling across the domain.

Table~\ref{tab:data_generation} summarizes the data generation settings.

\begin{table}[!ht]
\centering
\caption{Analytical data generation settings.}
\label{tab:data_generation}

\begin{tabular}{lccc}
\toprule
Function Type & Domain & Training Samples & Sampling Method\\
\midrule
Trigonometric & $[-2\pi,2\pi]$ & 300 & Uniform\\
Exponential & $[-2,2]$ & 300 & Uniform\\
Polynomial & $[-2,2]$ & 300 & Uniform\\
\bottomrule
\end{tabular}

\end{table}

An independent test set containing 200 uniformly sampled points over the same domain was used exclusively for evaluating approximation performance.

\subsection{Shallow Neural Network Experiments}

Single-hidden-layer feedforward neural networks were trained using three activation functions:

\begin{itemize}
\item ReLU
\item Hyperbolic tangent (tanh)
\item Sigmoid
\end{itemize}

The output layer employed a linear activation function. The number of hidden neurons was varied from 1 to 75. All networks were trained using the mean squared error loss function and stochastic gradient descent. Parameters were initialized randomly. For every function--activation pair, models were ranked according to numerically indistinguisable $R^2$

Table~\ref{tab:shallow_settings} summarizes the complete experimental configuration.

\begin{table}[!ht]

\centering
\caption{Shallow neural network configuration.}
\label{tab:shallow_settings}

\begin{tabular}{lc}
\toprule
Parameter & Value\\
\midrule
Architecture & Single hidden layer\\
Hidden neurons & 1--75\\
Activation functions & ReLU, tanh, Sigmoid\\
Output activation & Linear\\
Loss function & Mean squared error\\
Training algorithm & Stochastic gradient descent\\
Weight initialization & Random\\
Selection criterion & Top-$K$ models satisfying $\mathrm{NRMSE}<\varepsilon$\\
\bottomrule
\end{tabular}

\end{table}

\subsection{Multilayer Perceptron Experiments}

To investigate functional equivalence across architectures of varying depth and width, multilayer perceptrons containing one to five hidden layers were examined. Each hidden layer contained between one and five neurons, yielding a total of 3905 unique network architectures. All architectures were trained independently using identical optimization settings. To evaluate robustness under measurement uncertainty, additive white Gaussian noise corresponding to an SNR of 12.04~dB (25\% noise level) was introduced into the training data. For both the noise-free and noisy datasets, models were ranked according to the coefficient of determination ($R^2$), and the Top-$K$ equivalent models were selected for further analysis. Computational complexity was quantified using the number of floating-point operations (FLOPs) required for one forward pass. Table~\ref{tab:mlp_settings} summarizes the complete experimental configuration.

\begin{table}[!ht]

\centering
\caption{Multilayer perceptron configuration.}
\label{tab:mlp_settings}

\begin{tabular}{lc}
\toprule
Parameter & Value\\
\midrule
Hidden layers & 1--5\\
Neurons per layer & 1--5\\
Total architectures & 3905\\
Noise conditions & Noise-free and 12.04 dB SNR\\
Ranking metric & $R^2$\\
Complexity metric & FLOPs\\
\bottomrule
\end{tabular}

\end{table}

\newpage
\subsection{Geometric Analysis}

The local geometry of each trained model was characterized by numerically approximating the Hessian matrix around the optimal parameter vector using central finite differences.

The following quantities were extracted from every Hessian:

\begin{itemize}
\item Eigenvalue spectrum
\item Effective rank
\item Singular value ratio
\end{itemize}

Consequently, every neural network was characterized using

\begin{itemize}
\item approximation accuracy,
\item number of trainable parameters,
\item FLOPs,
\item effective rank,
\item singular value ratio.
\end{itemize}

\subsection{Reproducibility}

To ensure robustness with respect to random initialization, every experiment was repeated using multiple random seeds. The complete optimization and geometric analysis pipeline was independently executed for each initialization. Unless otherwise stated, reported results correspond to the sample mean and sample standard deviation computed across these independent runs.

\FloatBarrier

\newcommand{\ShallowFigure}[5]{
\begin{figure}[!ht]
\centering
\includegraphics[
width=\textwidth,
trim=0cm 13.5cm 0cm 18cm,
clip
]{#1}

\caption{
Analysis of the #2 function using the #3 activation function.
(a) Top-$K$ functionally equivalent approximations.
(b) Hessian ellipses illustrating geometric diversity.
(c) Distribution of equivalent networks with respect to hidden neurons and training $R^2$.
(d) Hessian eigenvalue spectra highlighting differences in parameter-space geometry despite functional equivalence.
}

\label{fig:#4_#5}

\end{figure}

\FloatBarrier
}

\section{Additional Results for Shallow Neural Networks}

This section presents additional visualizations for the shallow neural network experiments. For every target function, the Top-$K$ functionally equivalent networks are shown together with their Hessian geometry, architectural complexity, and eigenvalue spectra.
\newpage

\subsection{Sine Function}
\begin{figure}[!ht]
\centering

\includegraphics[
    width=\textwidth,
    trim=0cm 0cm 0cm 0cm,
    clip
]{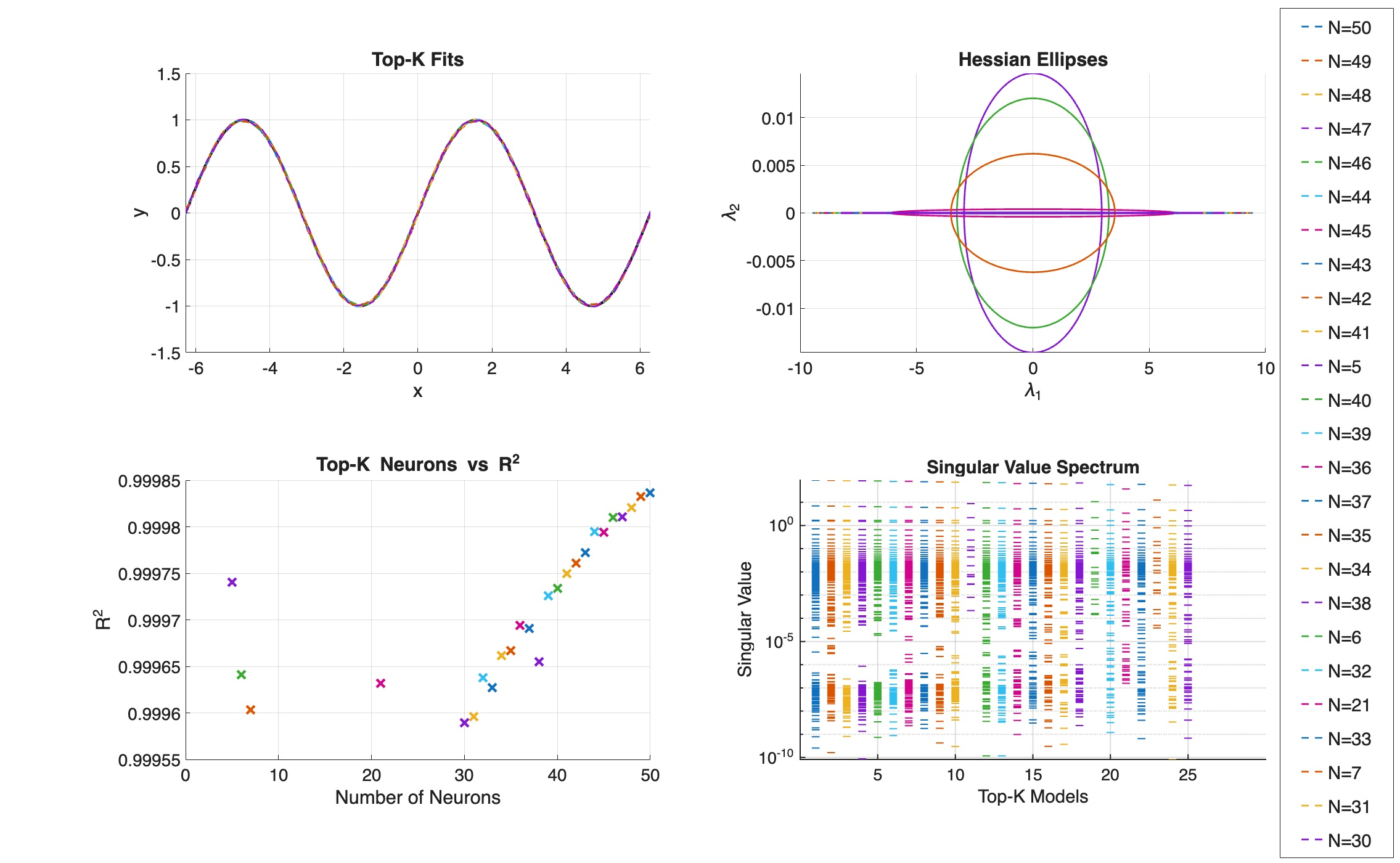}

\caption{
Analysis of the sin function using the tanh activation function. (a) Top-K functionally
equivalent approximations. (b) Hessian ellipses illustrating geometric diversity. (c) Distribution of equiva-
lent networks with respect to hidden neurons and training R2. (d) Hessian eigenvalue spectra highlighting
differences in parameter-space geometry despite functional equivalence}

\label{fig:sin_tanh}

\end{figure}

\FloatBarrier

\begin{figure}[!ht]
\centering

\includegraphics[
    width=\textwidth,
    trim=0cm 0cm 0cm 0cm,
    clip
]{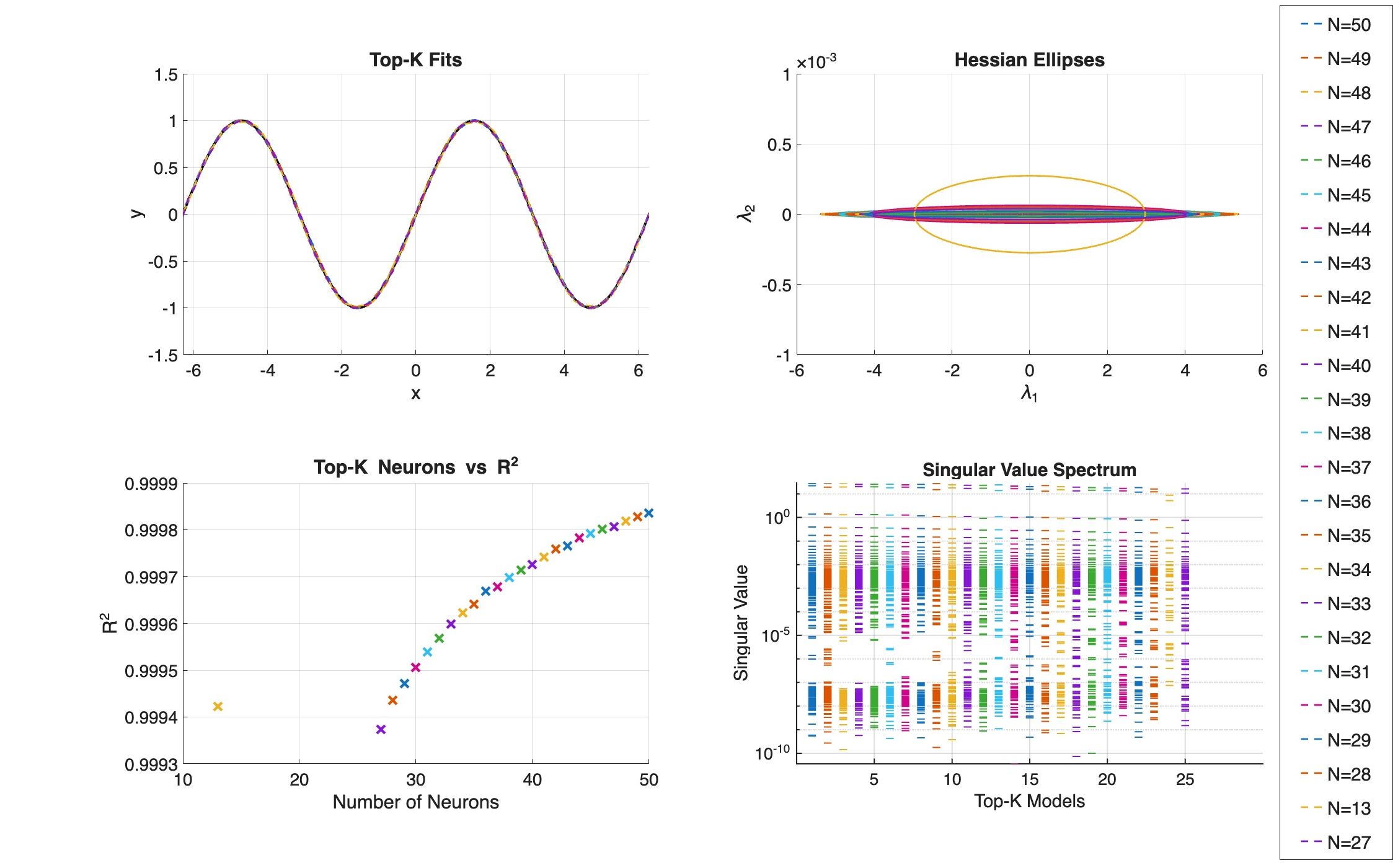}

\caption{
Analysis of the sin function using the sigmoid activation function. (a) Top-K functionally
equivalent approximations. (b) Hessian ellipses illustrating geometric diversity. (c) Distribution of equiva-
lent networks with respect to hidden neurons and training R2. (d) Hessian eigenvalue spectra highlighting
differences in parameter-space geometry despite functional equivalence}

\label{fig:sin_sigmoid}

\end{figure}

\FloatBarrier

\begin{figure}[!ht]
\centering

\includegraphics[
    width=\textwidth,
    trim=0cm 0cm 0cm 0cm,
    clip
]{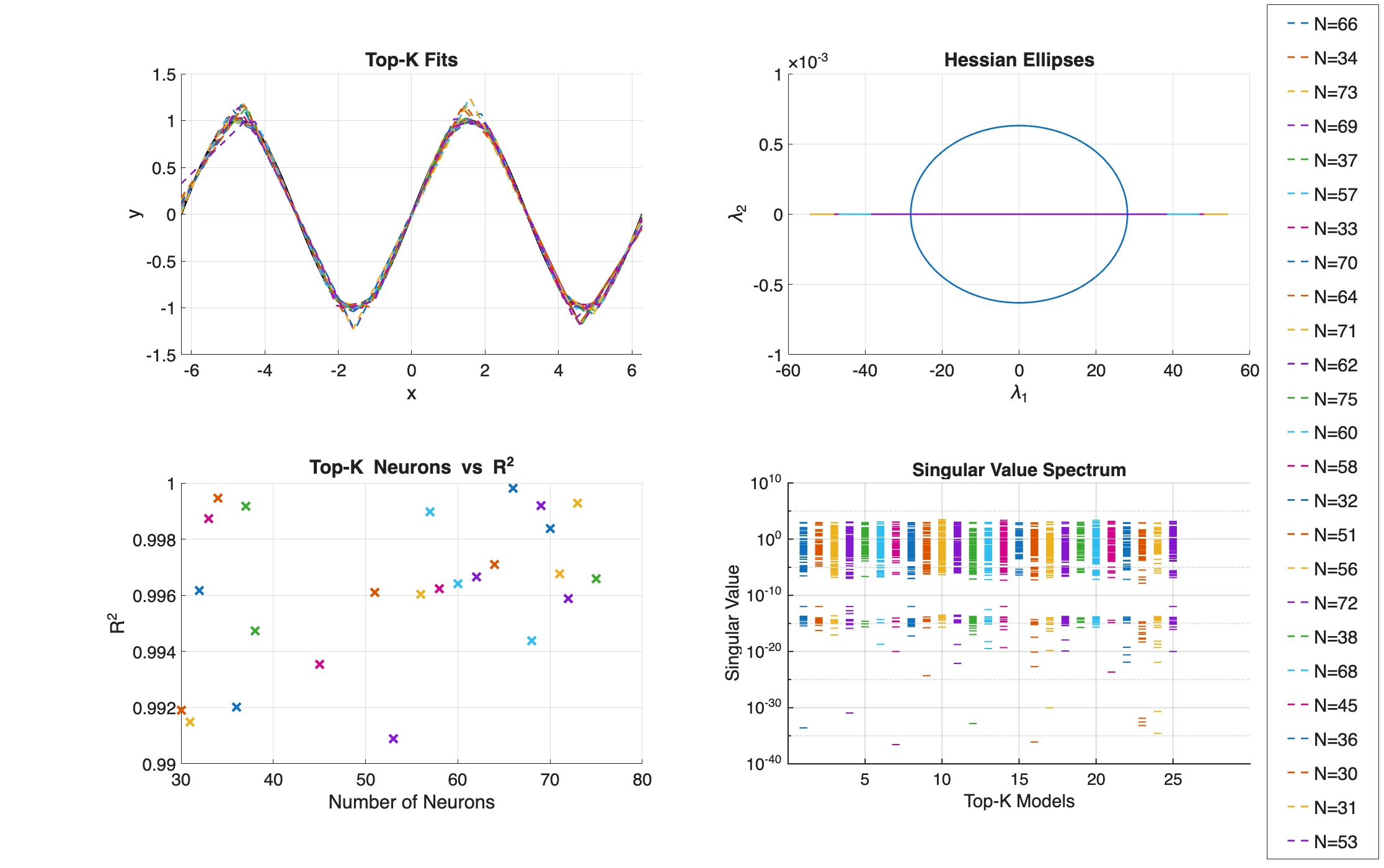}

\caption{
Analysis of the sin function using the RELU activation function. (a) Top-K functionally
equivalent approximations. (b) Hessian ellipses illustrating geometric diversity. (c) Distribution of equivalent networks with respect to hidden neurons and training R2. (d) Hessian eigenvalue spectra highlighting
differences in parameter-space geometry despite functional equivalence}

\label{fig:sin_relu}

\end{figure}

\FloatBarrier

\subsection{Cosine Function}
\begin{figure}[!ht]
\centering

\includegraphics[
    width=\textwidth,
    trim=0cm 0cm 0cm 0cm,
    clip
]{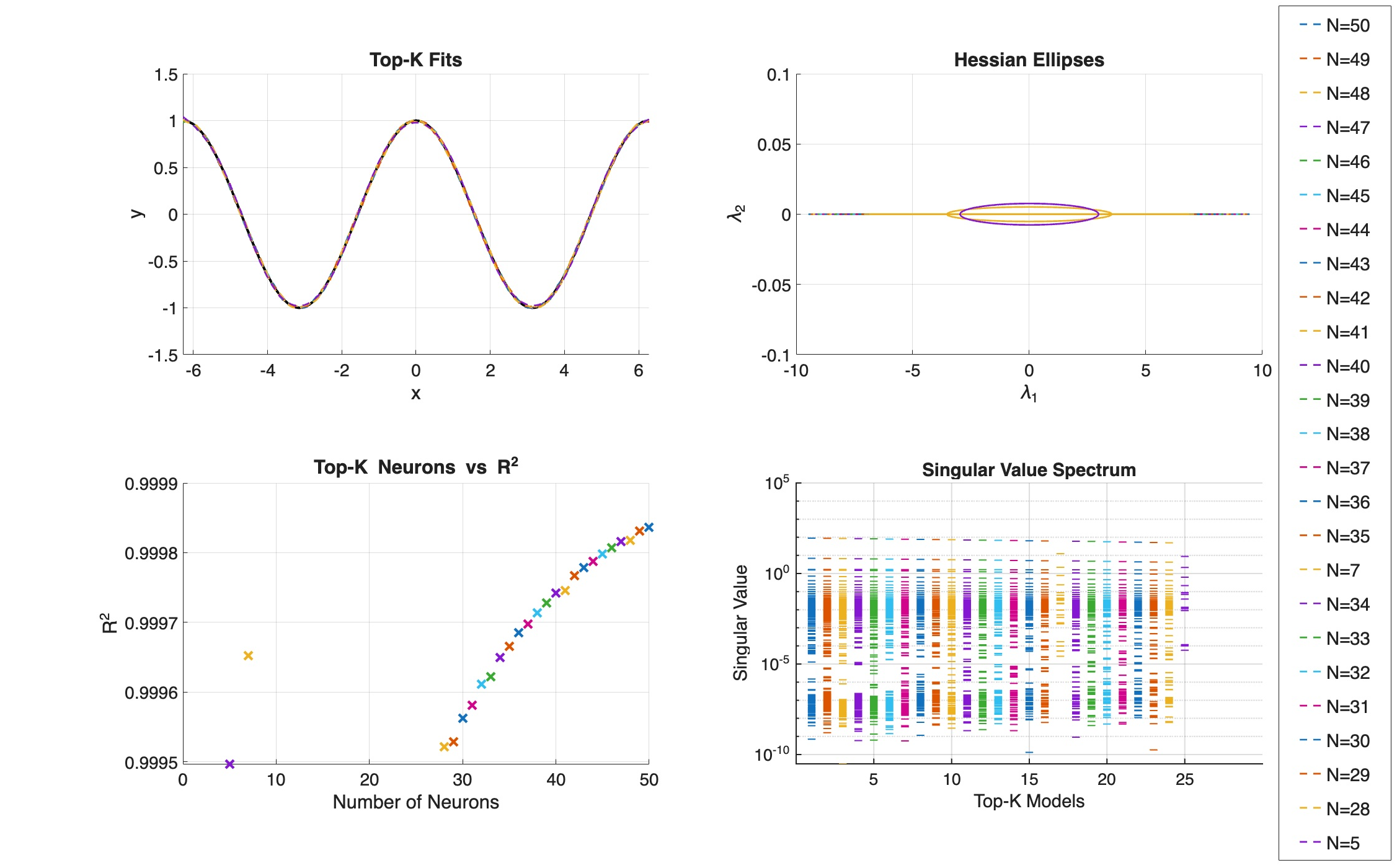}

\caption{
Analysis of the cosin function using the tanh activation function. (a) Top-K functionally
equivalent approximations. (b) Hessian ellipses illustrating geometric diversity. (c) Distribution of equiva-
lent networks with respect to hidden neurons and training R2. (d) Hessian eigenvalue spectra highlighting
differences in parameter-space geometry despite functional equivalence}

\label{fig:cos_tanh}

\end{figure}

\FloatBarrier

\begin{figure}[!ht]
\centering

\includegraphics[
    width=\textwidth,
    trim=0cm 0cm 0cm 0cm,
    clip
]{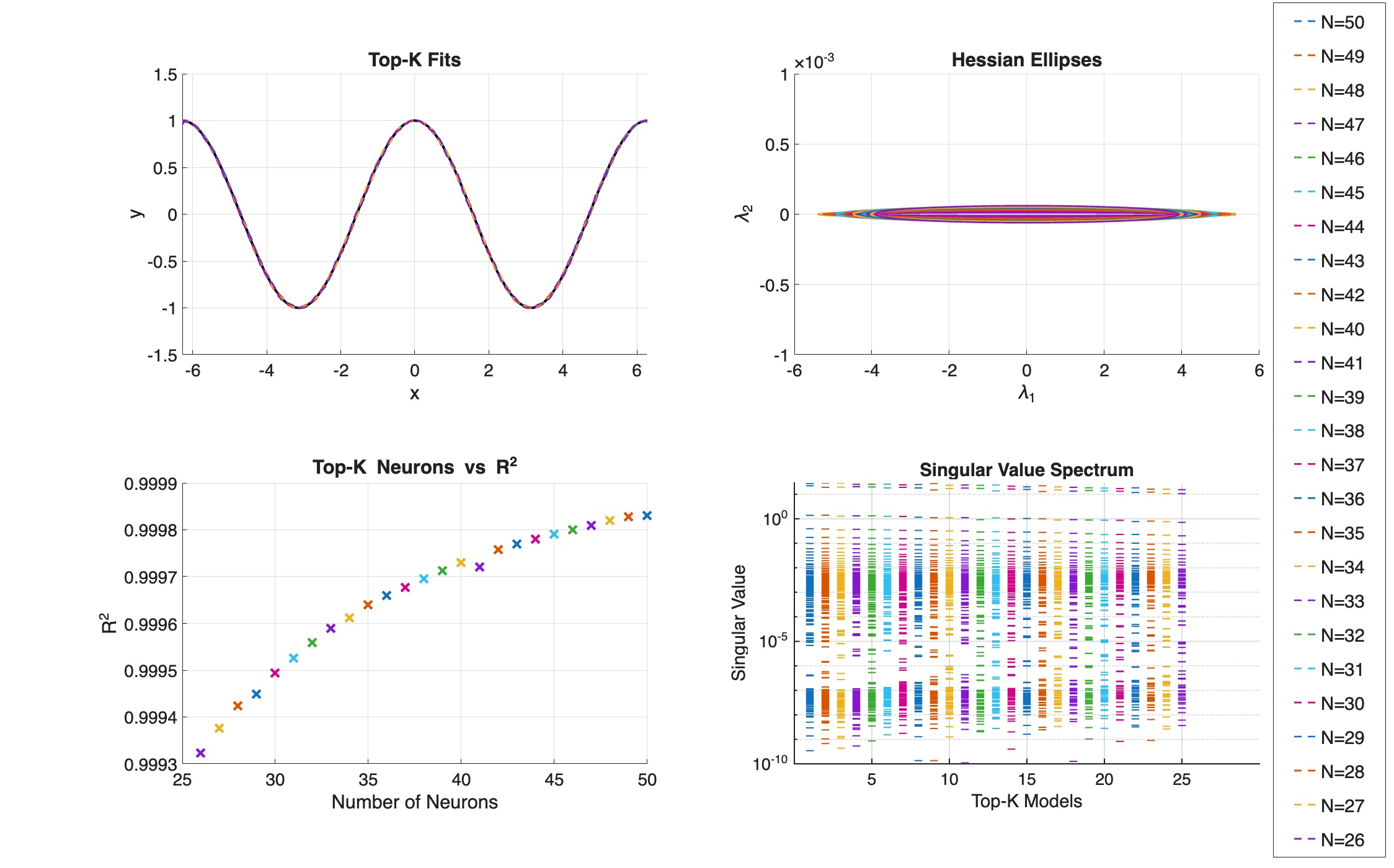}

\caption{
Analysis of the cosin function using the sigmoid activation function. (a) Top-K functionally
equivalent approximations. (b) Hessian ellipses illustrating geometric diversity. (c) Distribution of equiva-
lent networks with respect to hidden neurons and training R2. (d) Hessian eigenvalue spectra highlighting
differences in parameter-space geometry despite functional equivalence}

\label{fig:cos_sigmoid}

\end{figure}

\FloatBarrier

\begin{figure}[!ht]
\centering

\includegraphics[
    width=\textwidth,
    trim=0cm 0cm 0cm 0cm,
    clip
]{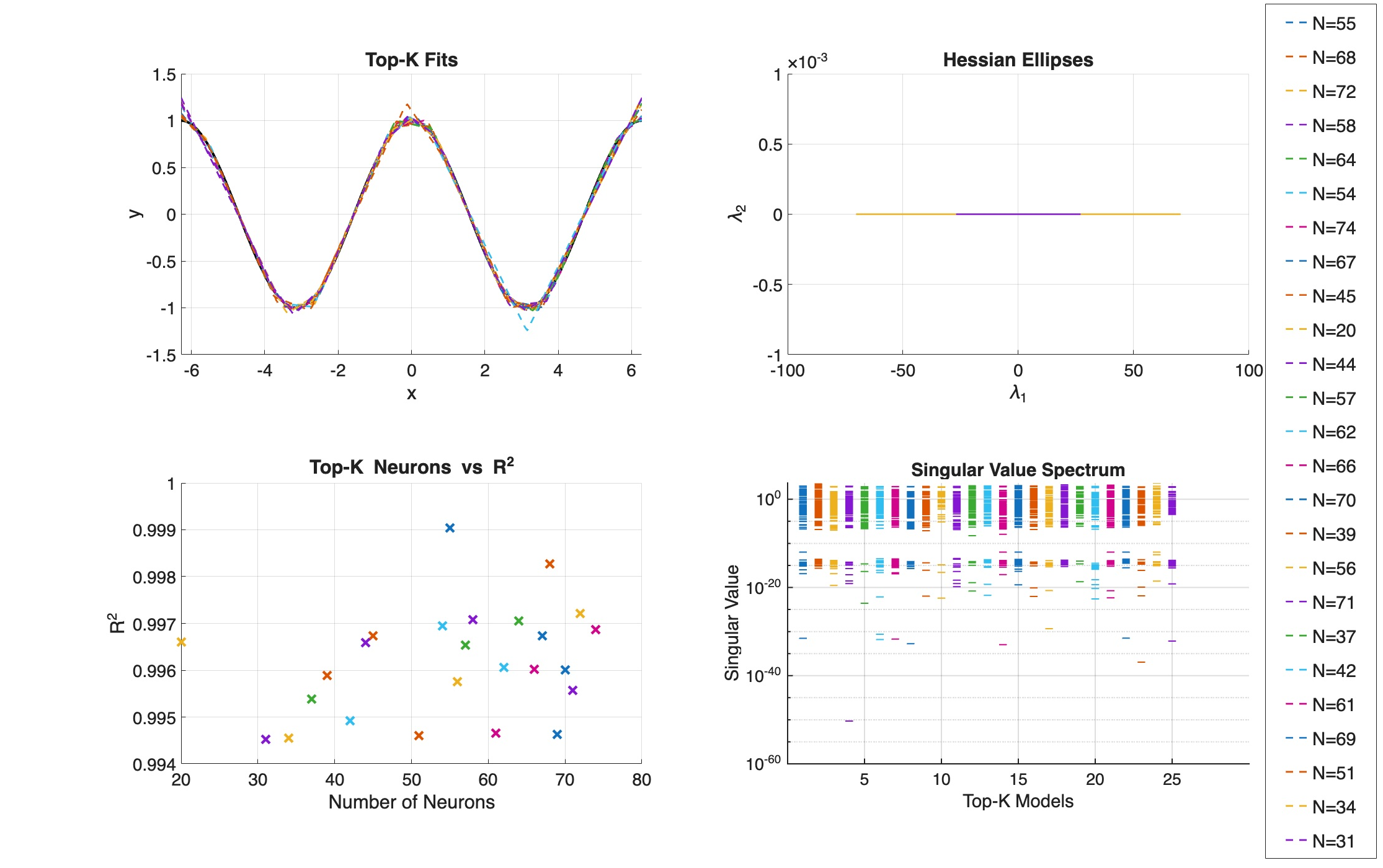}

\caption{
Analysis of the cosin function using the RELU activation function. (a) Top-K functionally
equivalent approximations. (b) Hessian ellipses illustrating geometric diversity. (c) Distribution of equivalent networks with respect to hidden neurons and training R2. (d) Hessian eigenvalue spectra highlighting
differences in parameter-space geometry despite functional equivalence}

\label{fig:cos_relu}

\end{figure}

\FloatBarrier

\subsection{Exponential Function}
\begin{figure}[!ht]
\centering

\includegraphics[
    width=\textwidth,
    trim=0cm 0cm 0cm 0cm,
    clip
]{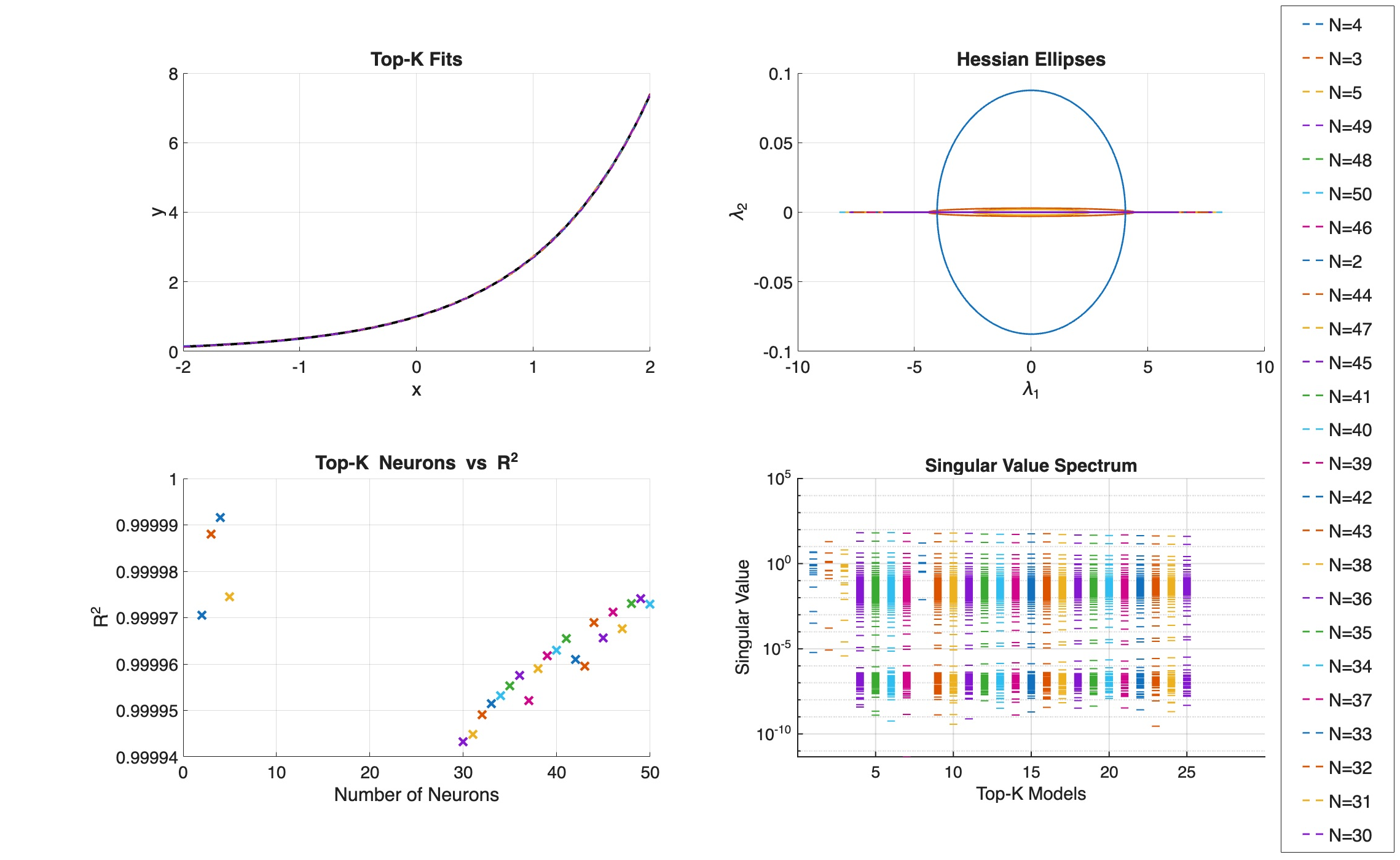}

\caption{
Analysis of the exponential function using the tanh activation function. (a) Top-K functionally
equivalent approximations. (b) Hessian ellipses illustrating geometric diversity. (c) Distribution of equiva-
lent networks with respect to hidden neurons and training R2. (d) Hessian eigenvalue spectra highlighting
differences in parameter-space geometry despite functional equivalence}

\label{fig:exp_tanh}

\end{figure}

\FloatBarrier

\begin{figure}[!ht]
\centering

\includegraphics[
    width=\textwidth,
    trim=0cm 0cm 0cm 0cm,
    clip
]{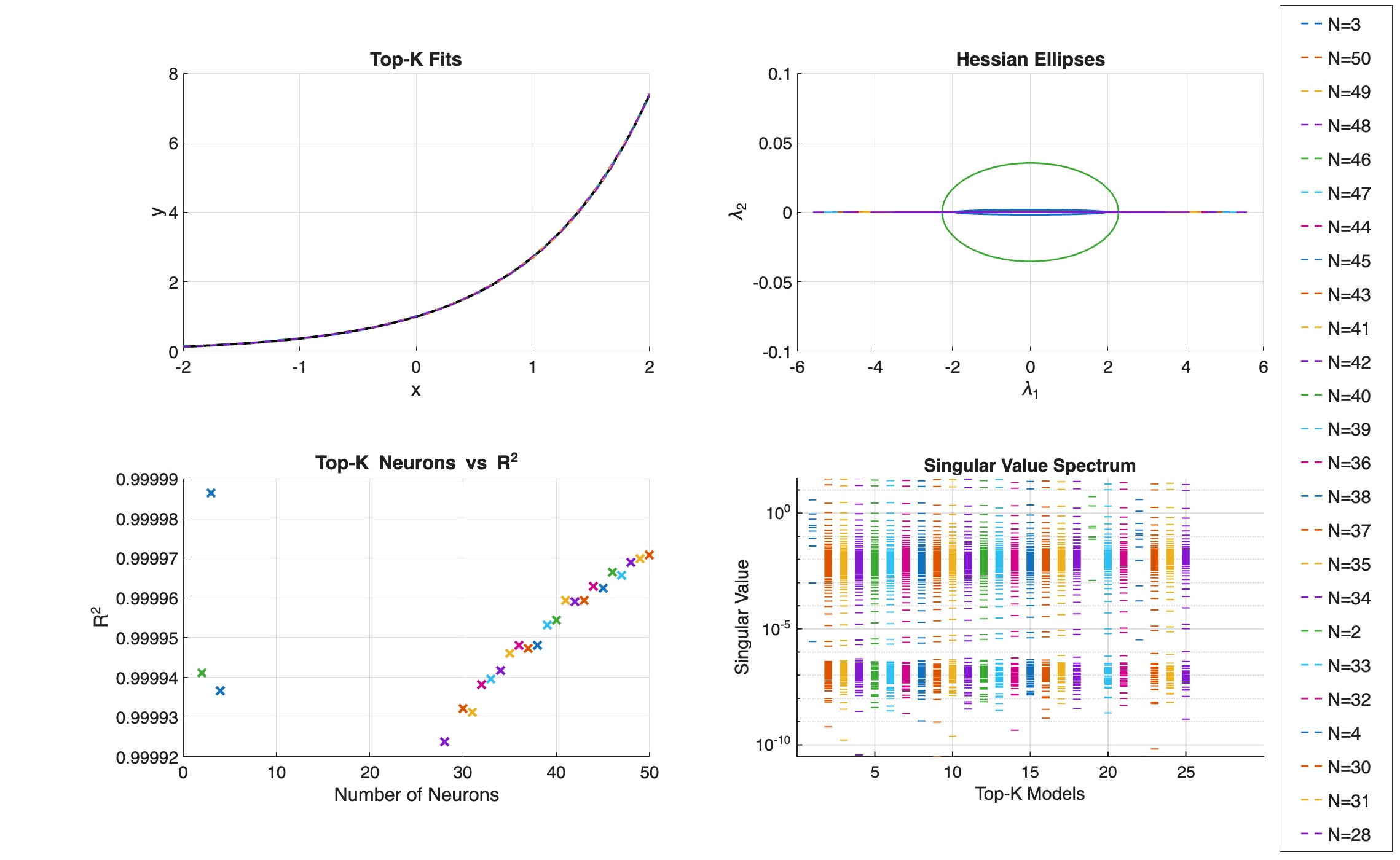}

\caption{
Analysis of the exponential function using the sigmoid activation function. (a) Top-K functionally
equivalent approximations. (b) Hessian ellipses illustrating geometric diversity. (c) Distribution of equiva-
lent networks with respect to hidden neurons and training R2. (d) Hessian eigenvalue spectra highlighting
differences in parameter-space geometry despite functional equivalence}

\label{fig:exp_sigmoid}

\end{figure}

\FloatBarrier

\begin{figure}[!ht]
\centering

\includegraphics[
    width=\textwidth,
    trim=0cm 0cm 0cm 0cm,
    clip
]{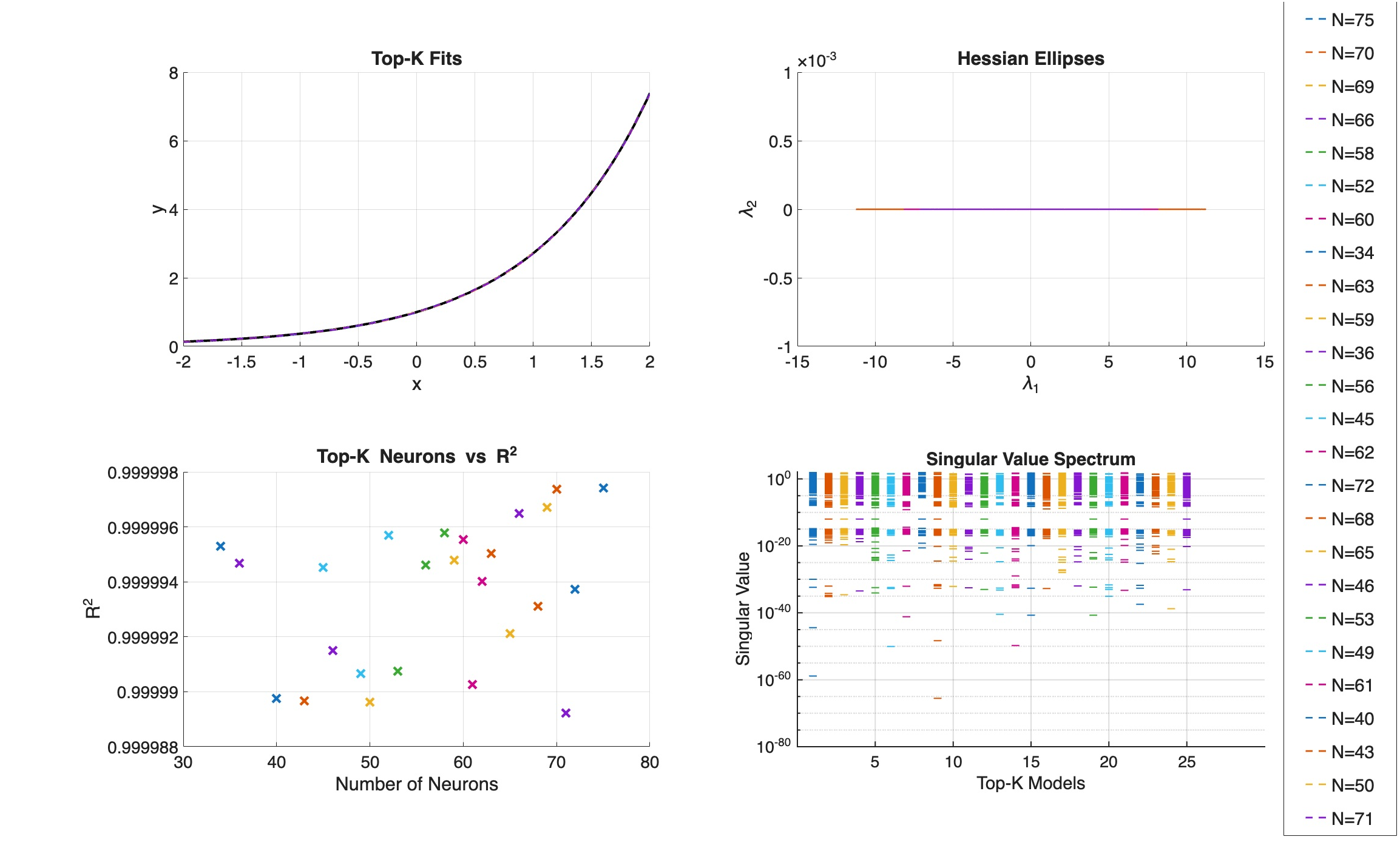}

\caption{
Analysis of the exponential function using the RELU activation function. (a) Top-K functionally
equivalent approximations. (b) Hessian ellipses illustrating geometric diversity. (c) Distribution of equivalent networks with respect to hidden neurons and training R2. (d) Hessian eigenvalue spectra highlighting
differences in parameter-space geometry despite functional equivalence}

\label{fig:exp_relu}

\end{figure}

\FloatBarrier

\subsection{Quadratic Function}
\begin{figure}[!ht]
\centering

\includegraphics[
    width=\textwidth,
    trim=0cm 0cm 0cm 0cm,
    clip
]{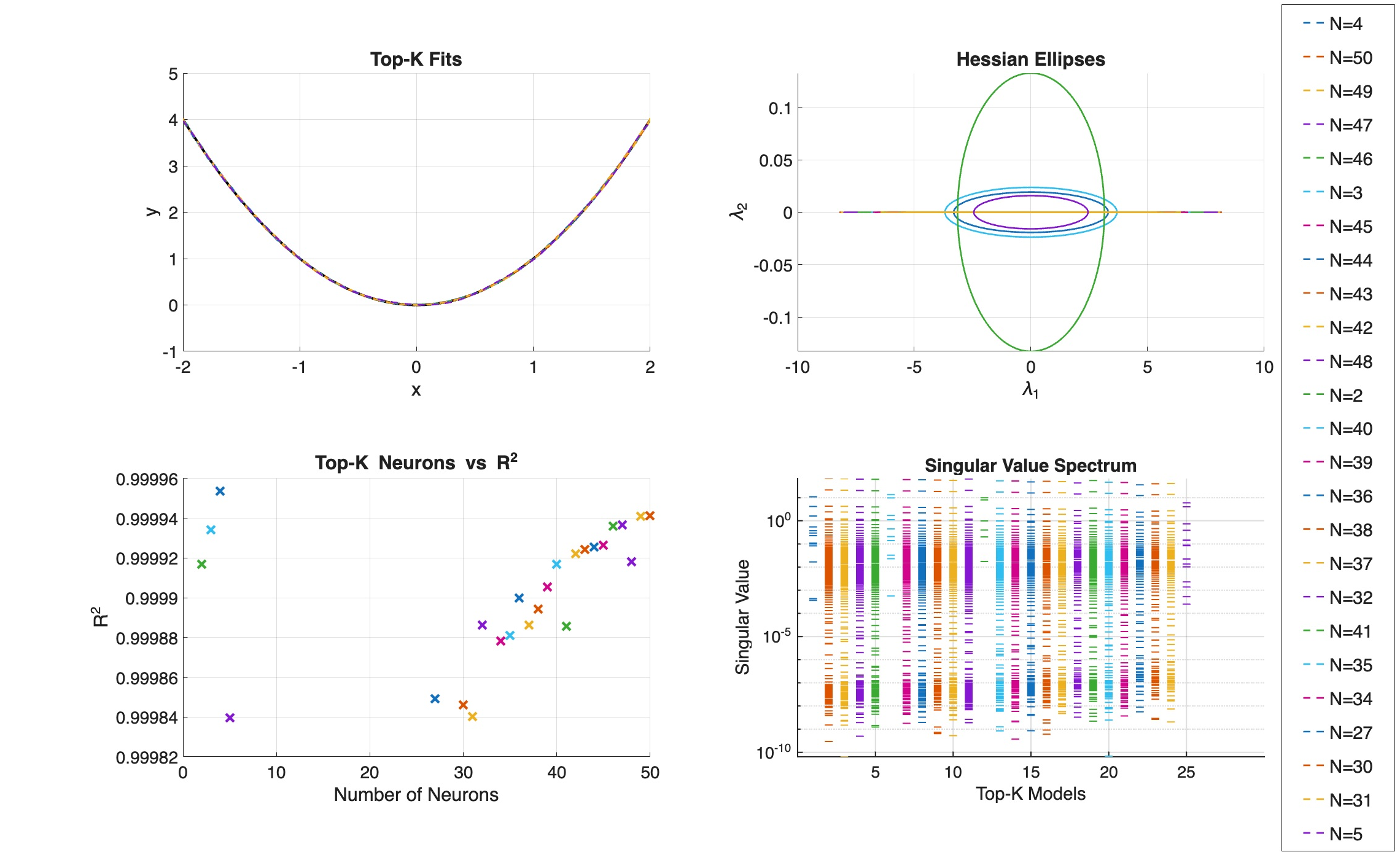}

\caption{
Analysis of the quadratic function using the tanh activation function. (a) Top-K functionally
equivalent approximations. (b) Hessian ellipses illustrating geometric diversity. (c) Distribution of equiva-
lent networks with respect to hidden neurons and training R2. (d) Hessian eigenvalue spectra highlighting
differences in parameter-space geometry despite functional equivalence}

\label{fig:square_tanh}

\end{figure}

\FloatBarrier

\begin{figure}[!ht]
\centering

\includegraphics[
    width=\textwidth,
    trim=0cm 0cm 0cm 0cm,
    clip
]{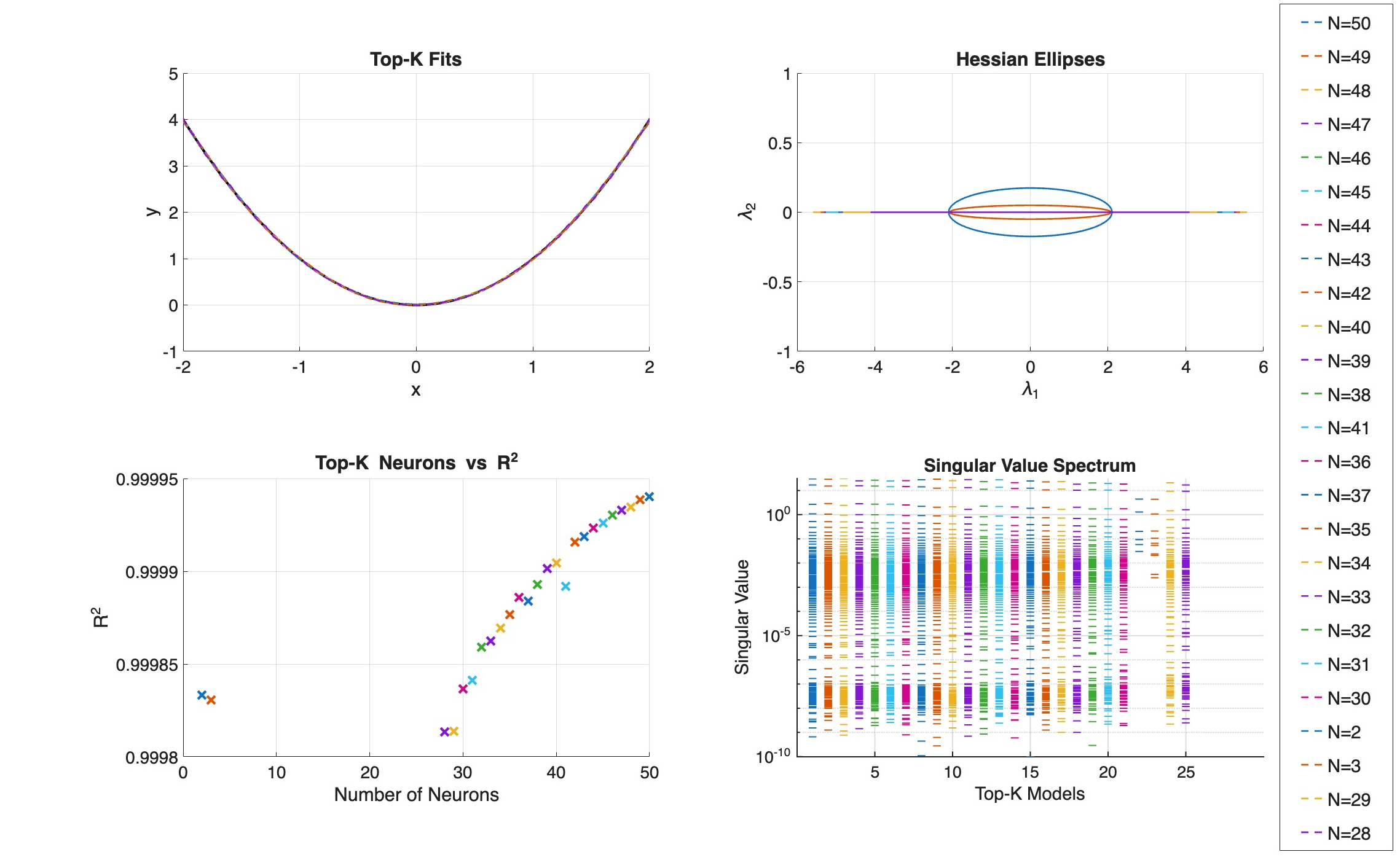}

\caption{
Analysis of the quadratic function using the sigmoid activation function. (a) Top-K functionally
equivalent approximations. (b) Hessian ellipses illustrating geometric diversity. (c) Distribution of equiva-
lent networks with respect to hidden neurons and training R2. (d) Hessian eigenvalue spectra highlighting
differences in parameter-space geometry despite functional equivalence}

\label{fig:square_sigmoid}

\end{figure}

\FloatBarrier

\begin{figure}[!ht]
\centering

\includegraphics[
    width=\textwidth,
    trim=0cm 0cm 0cm 0cm,
    clip
]{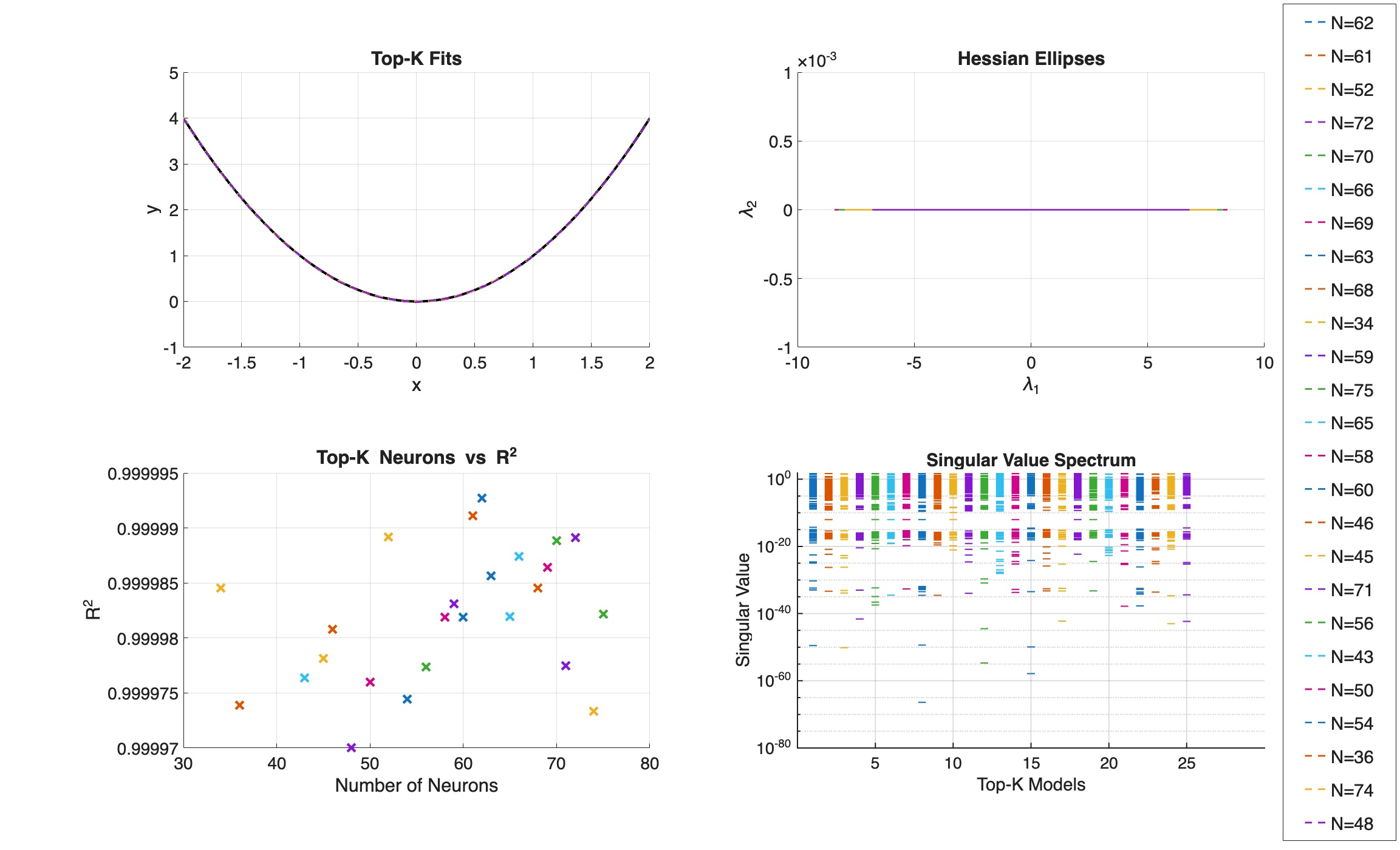}

\caption{
Analysis of the quadratic function using the RELU activation function. (a) Top-K functionally
equivalent approximations. (b) Hessian ellipses illustrating geometric diversity. (c) Distribution of equivalent networks with respect to hidden neurons and training R2. (d) Hessian eigenvalue spectra highlighting
differences in parameter-space geometry despite functional equivalence}

\label{fig:square_relu}

\end{figure}

\FloatBarrier

\subsection{Sine Function}
\begin{figure}[!ht]
\centering

\includegraphics[
    width=\textwidth,
    trim=0cm 0cm 0cm 0cm,
    clip
]{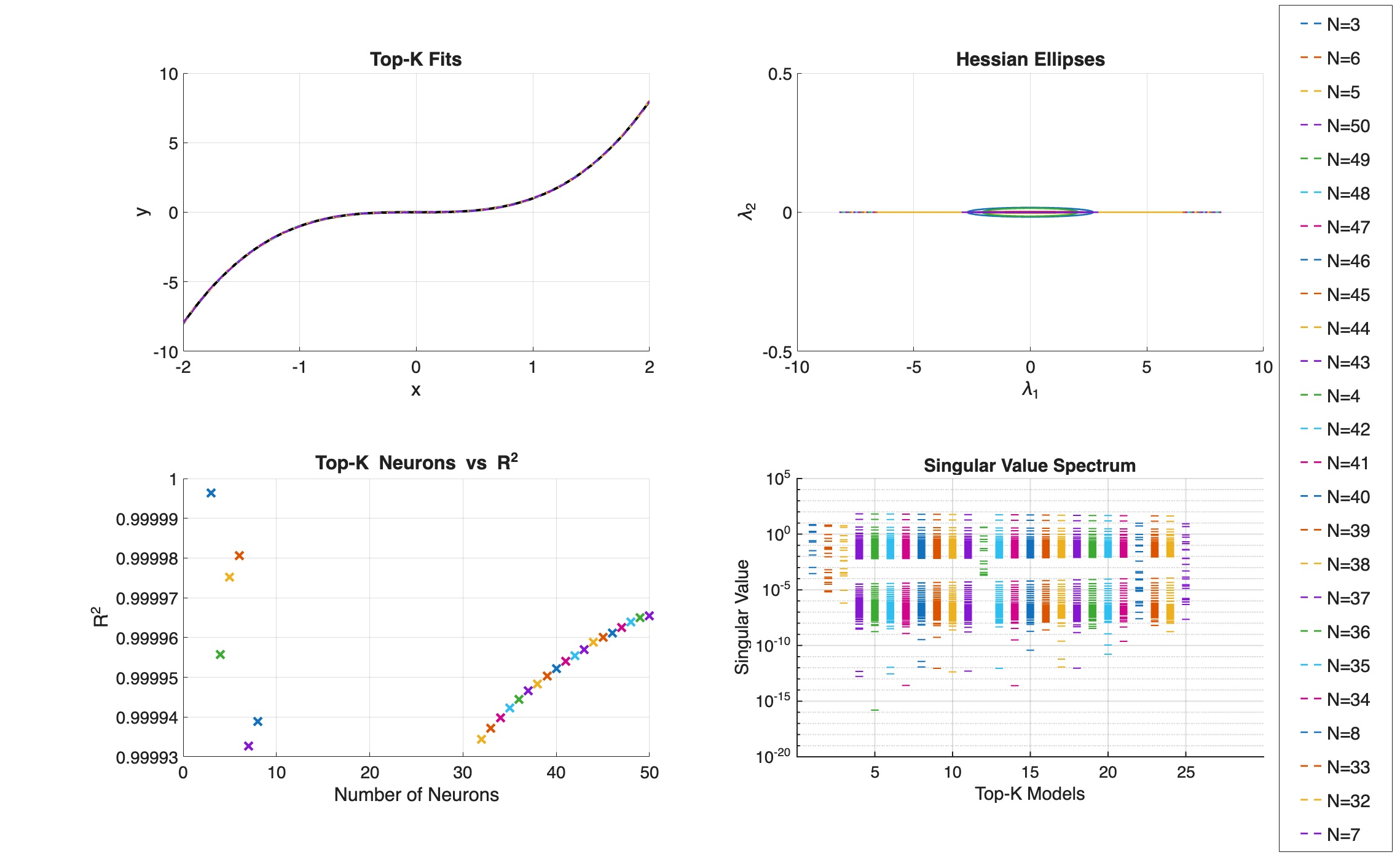}

\caption{
Analysis of the cubic function using the tanh activation function. (a) Top-K functionally
equivalent approximations. (b) Hessian ellipses illustrating geometric diversity. (c) Distribution of equiva-
lent networks with respect to hidden neurons and training R2. (d) Hessian eigenvalue spectra highlighting
differences in parameter-space geometry despite functional equivalence}

\label{fig:cube_tanh}

\end{figure}

\FloatBarrier

\begin{figure}[!ht]
\centering

\includegraphics[
    width=\textwidth,
    trim=0cm 0cm 0cm 0cm,
    clip
]{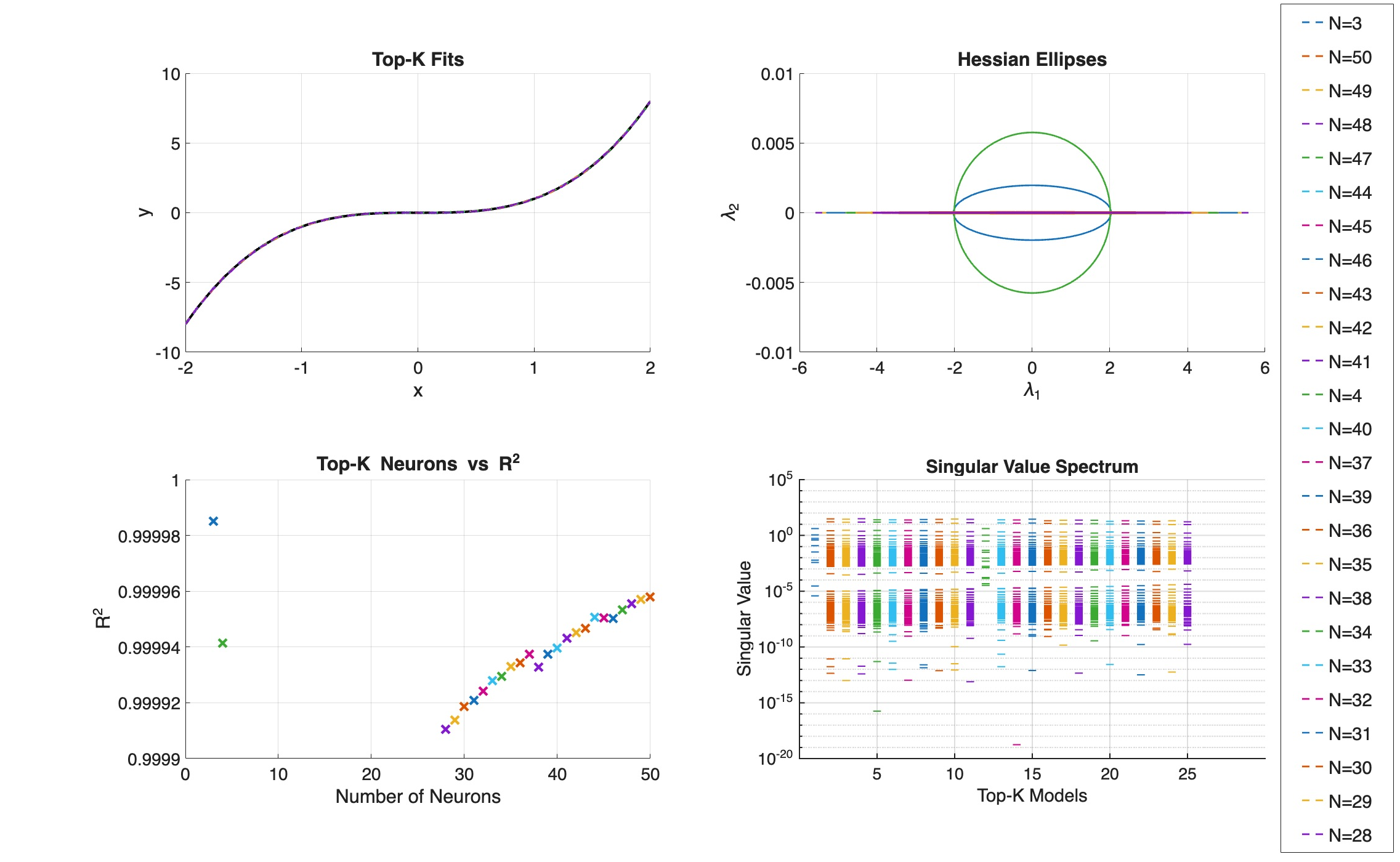}

\caption{
Analysis of the cubic function using the sigmoid activation function. (a) Top-K functionally
equivalent approximations. (b) Hessian ellipses illustrating geometric diversity. (c) Distribution of equiva-
lent networks with respect to hidden neurons and training R2. (d) Hessian eigenvalue spectra highlighting
differences in parameter-space geometry despite functional equivalence}

\label{fig:cube_sigmoid}

\end{figure}

\FloatBarrier

\begin{figure}[!ht]
\centering

\includegraphics[
    width=\textwidth,
    trim=0cm 0cm 0cm 0cm,
    clip
]{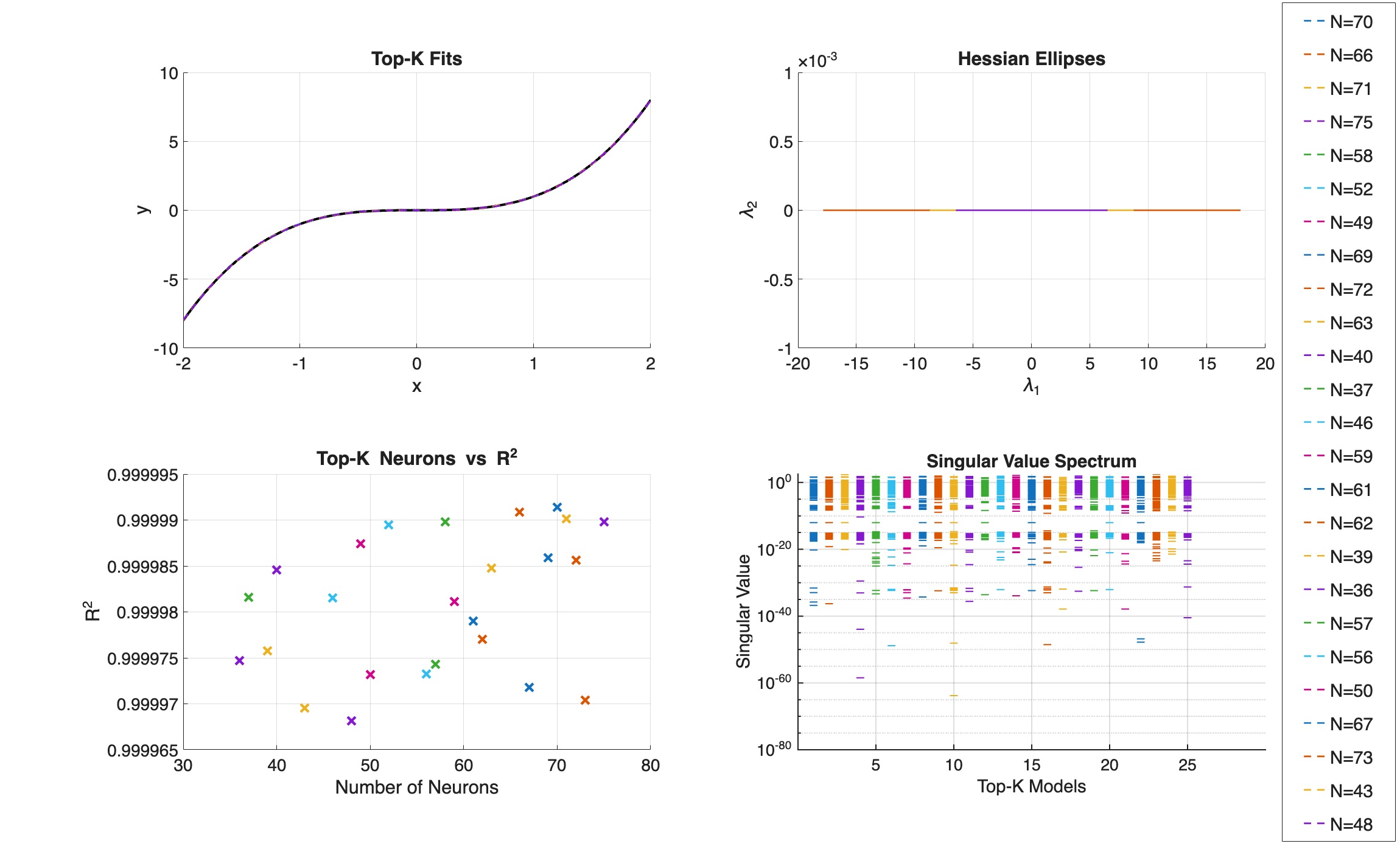}

\caption{
Analysis of the cubic function using the RELU activation function. (a) Top-K functionally
equivalent approximations. (b) Hessian ellipses illustrating geometric diversity. (c) Distribution of equivalent networks with respect to hidden neurons and training R2. (d) Hessian eigenvalue spectra highlighting
differences in parameter-space geometry despite functional equivalence}

\label{fig:cube_relu}

\end{figure}

\FloatBarrier

\begin{figure}[!ht]

\centering

\begin{tabular}{ccc}

\includegraphics[width=.31\textwidth]{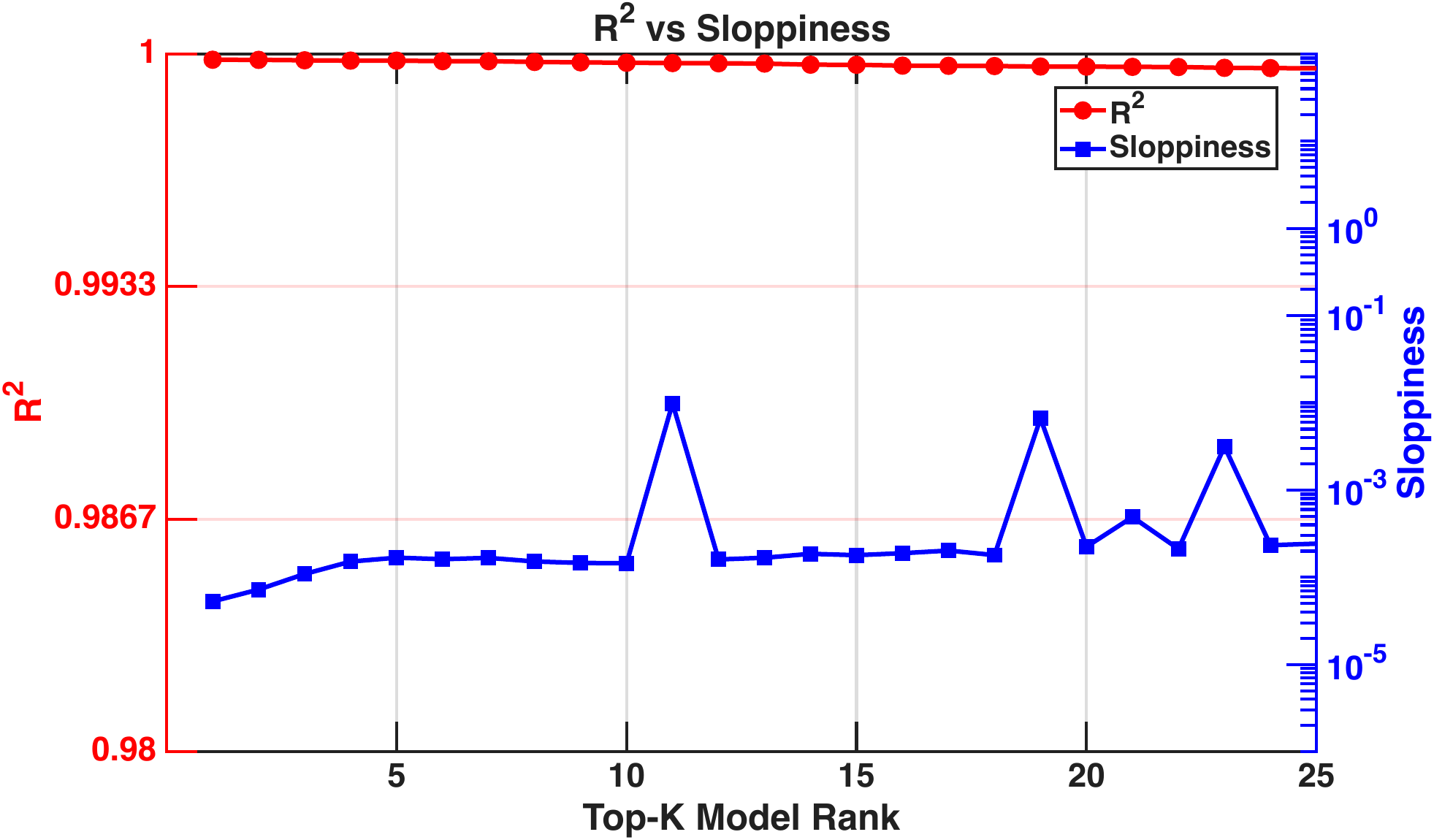}
&
\includegraphics[width=.31\textwidth]{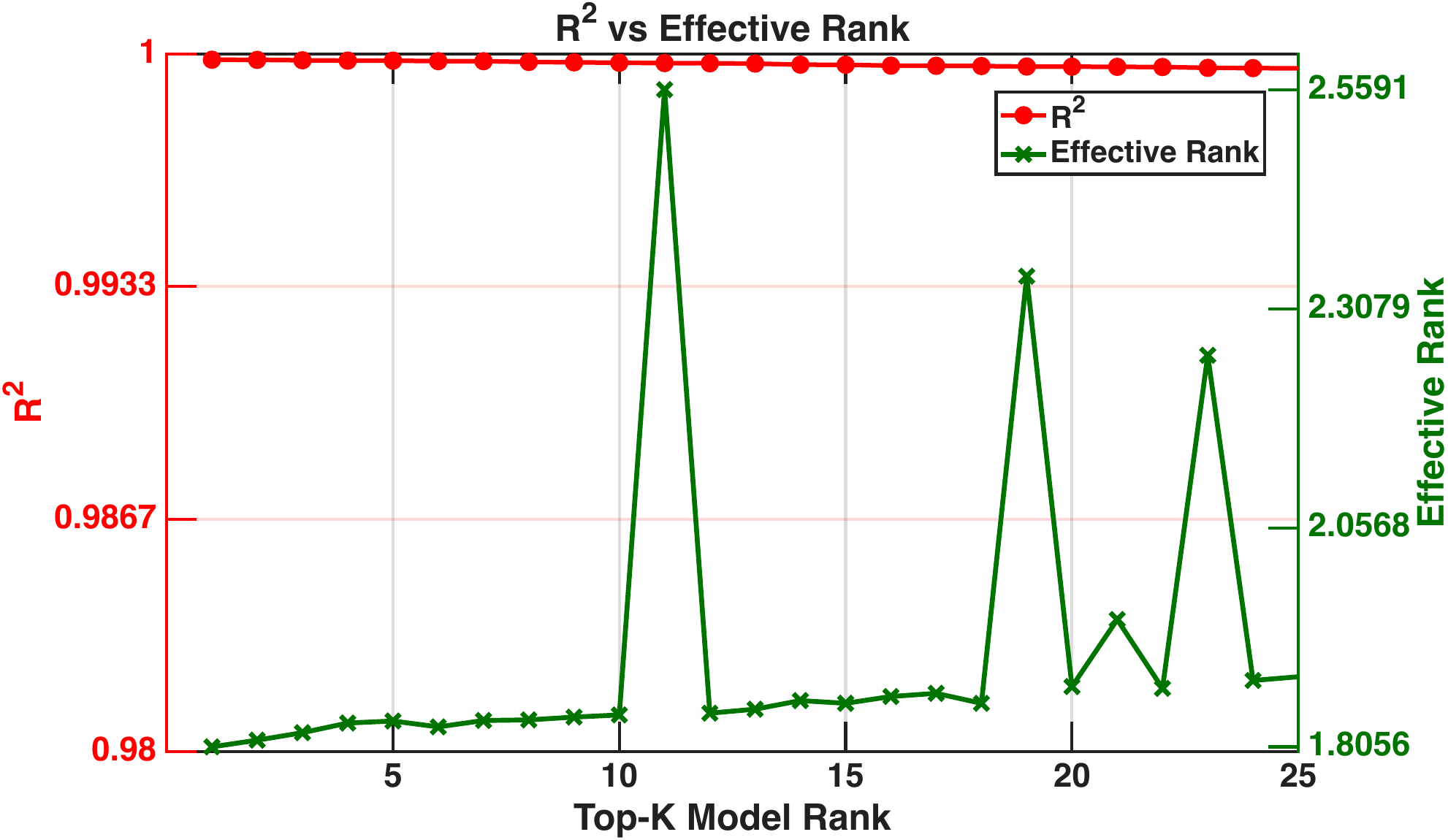}
&
\includegraphics[width=.31\textwidth]{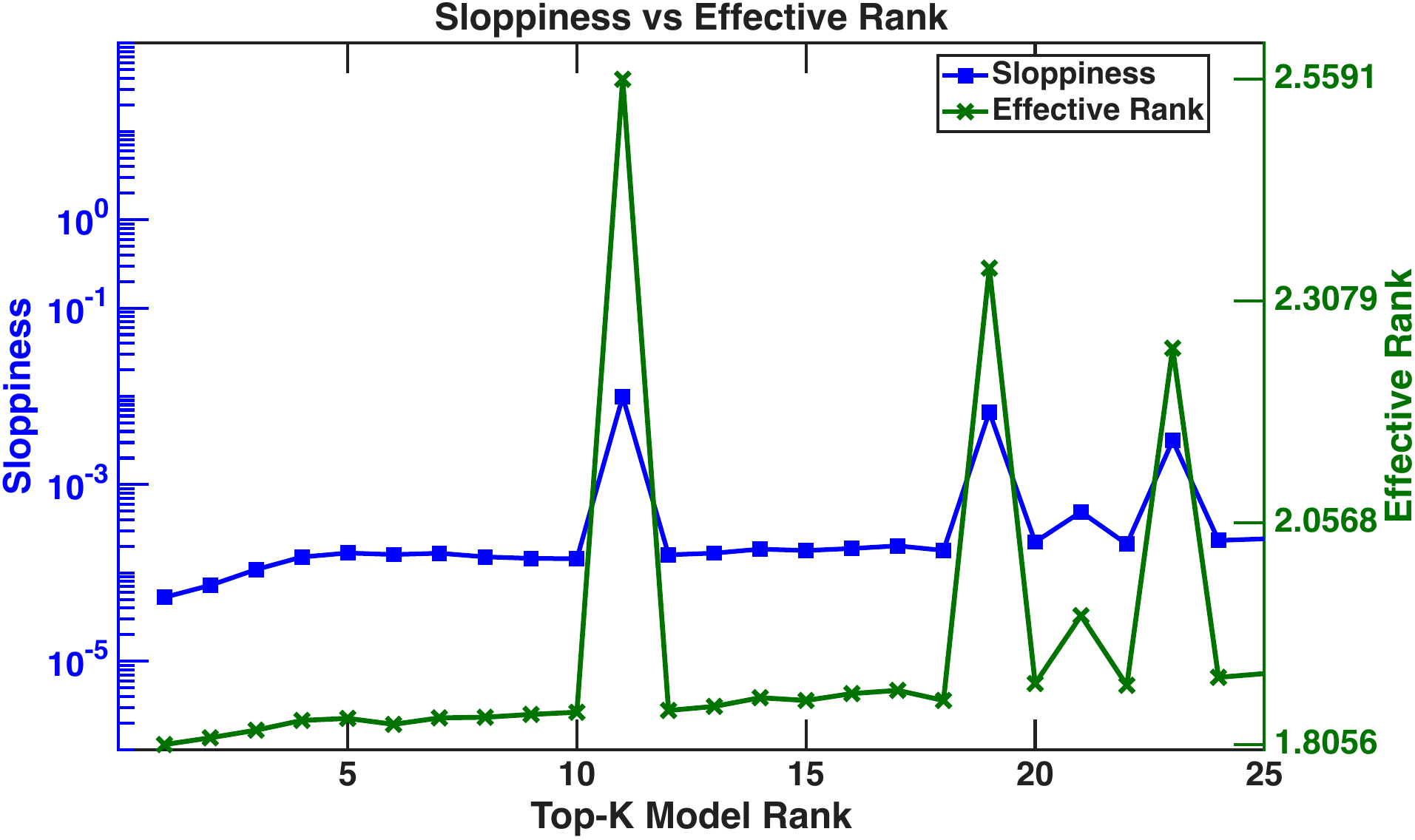}

\\[2mm]

\includegraphics[width=.31\textwidth]{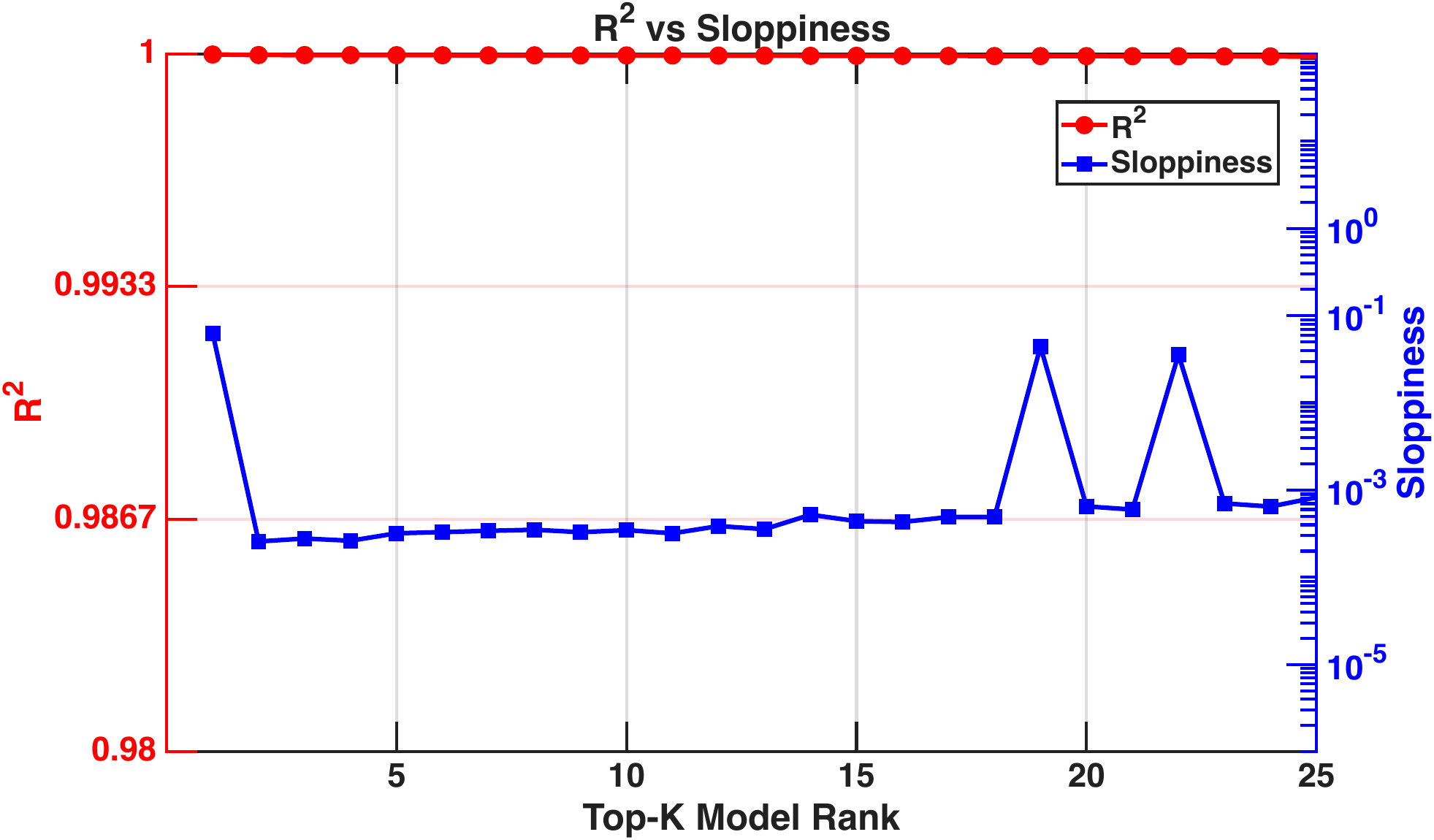}
&
\includegraphics[width=.31\textwidth]{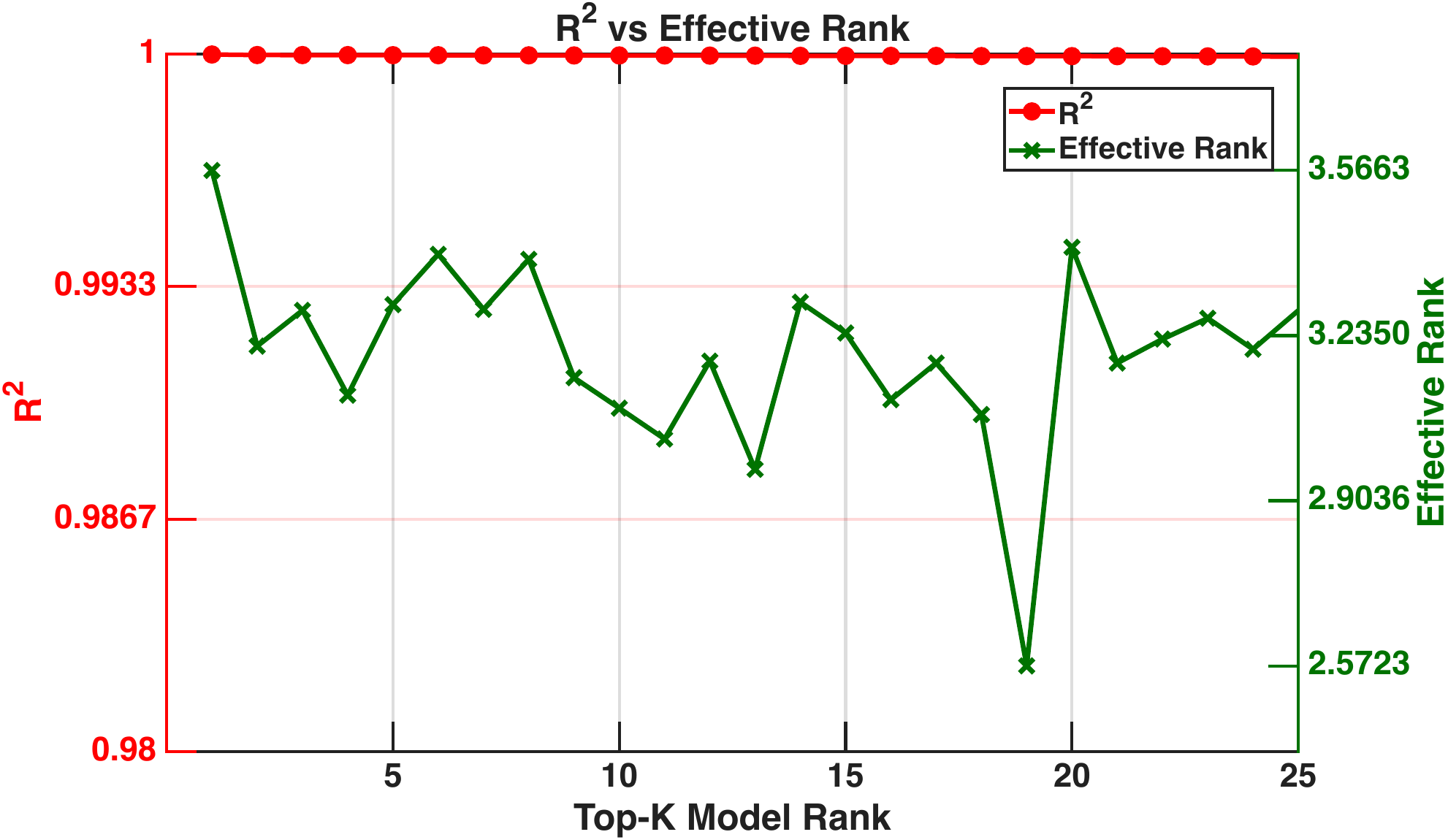}
&
\includegraphics[width=.31\textwidth]{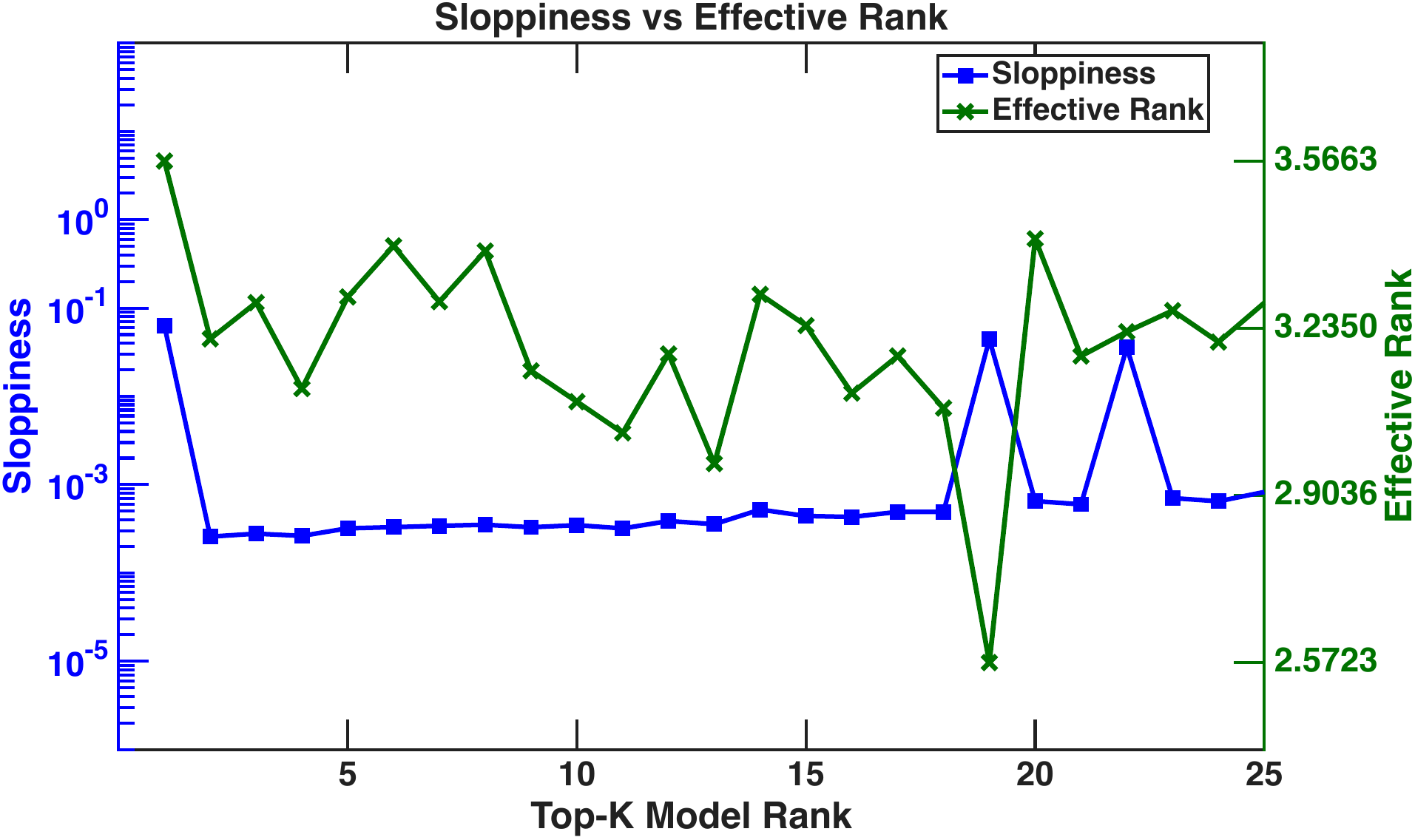}

\\[2mm]

\includegraphics[width=.31\textwidth]{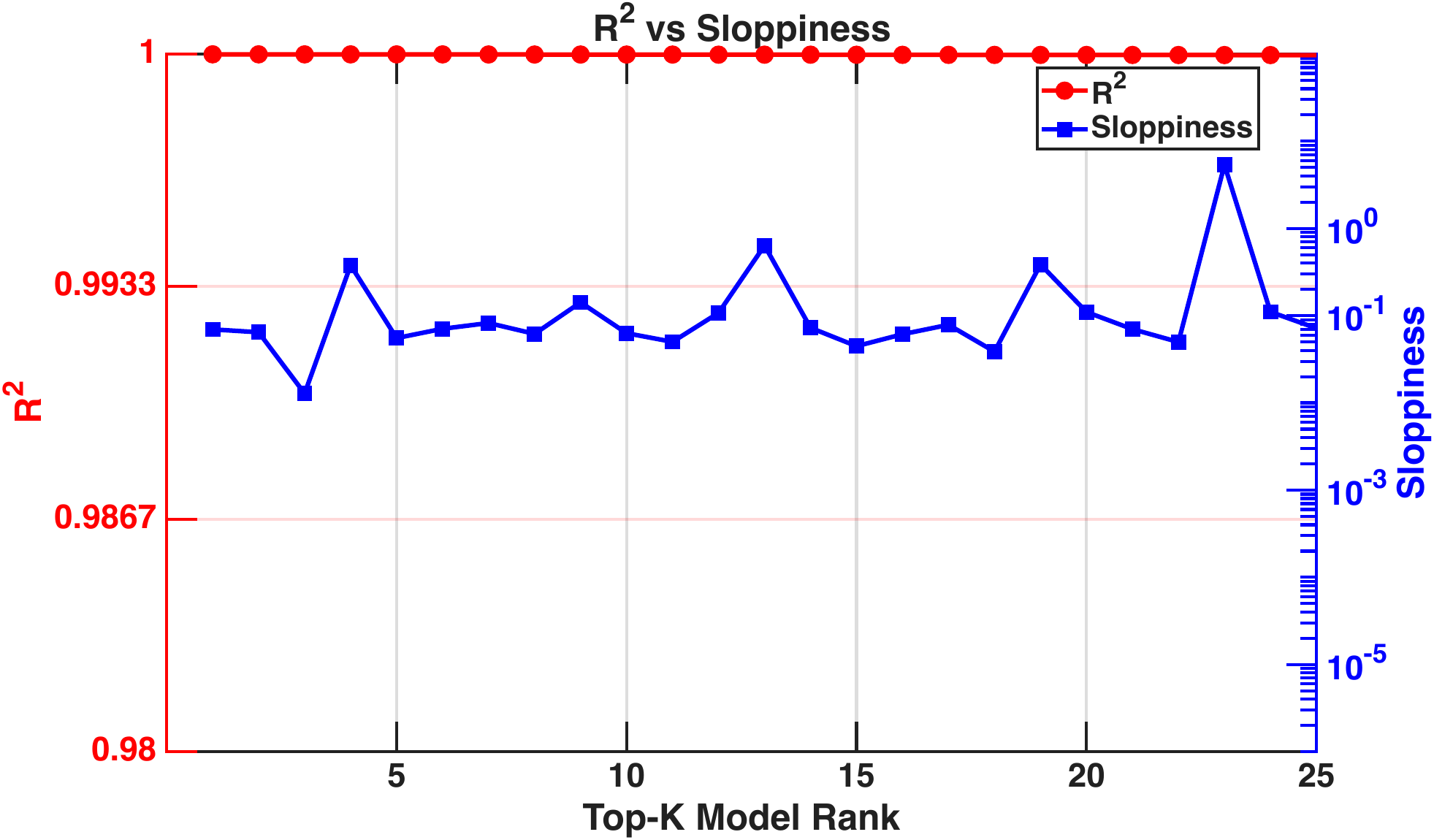}
&
\includegraphics[width=.31\textwidth]{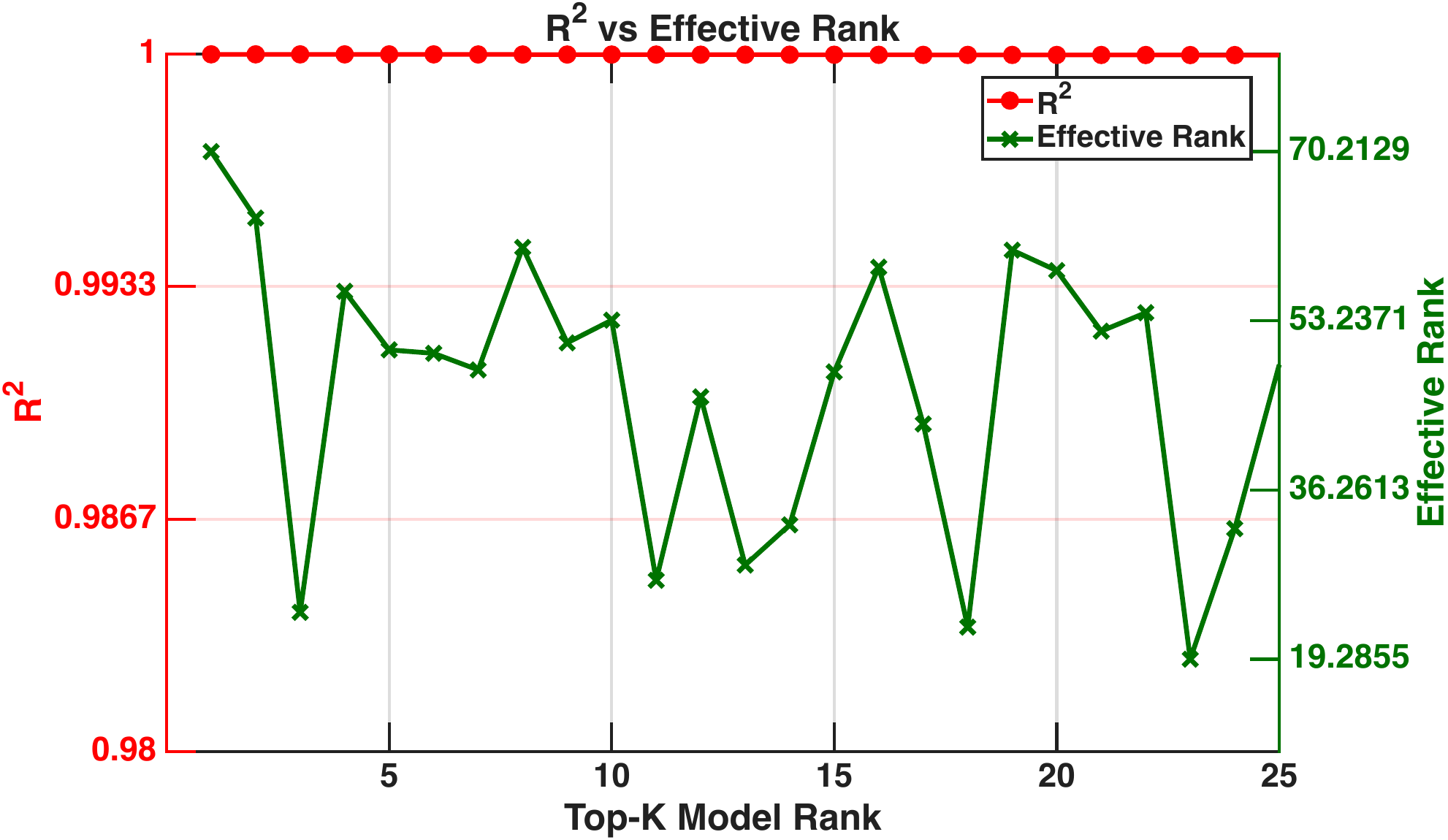}
&
\includegraphics[width=.31\textwidth]{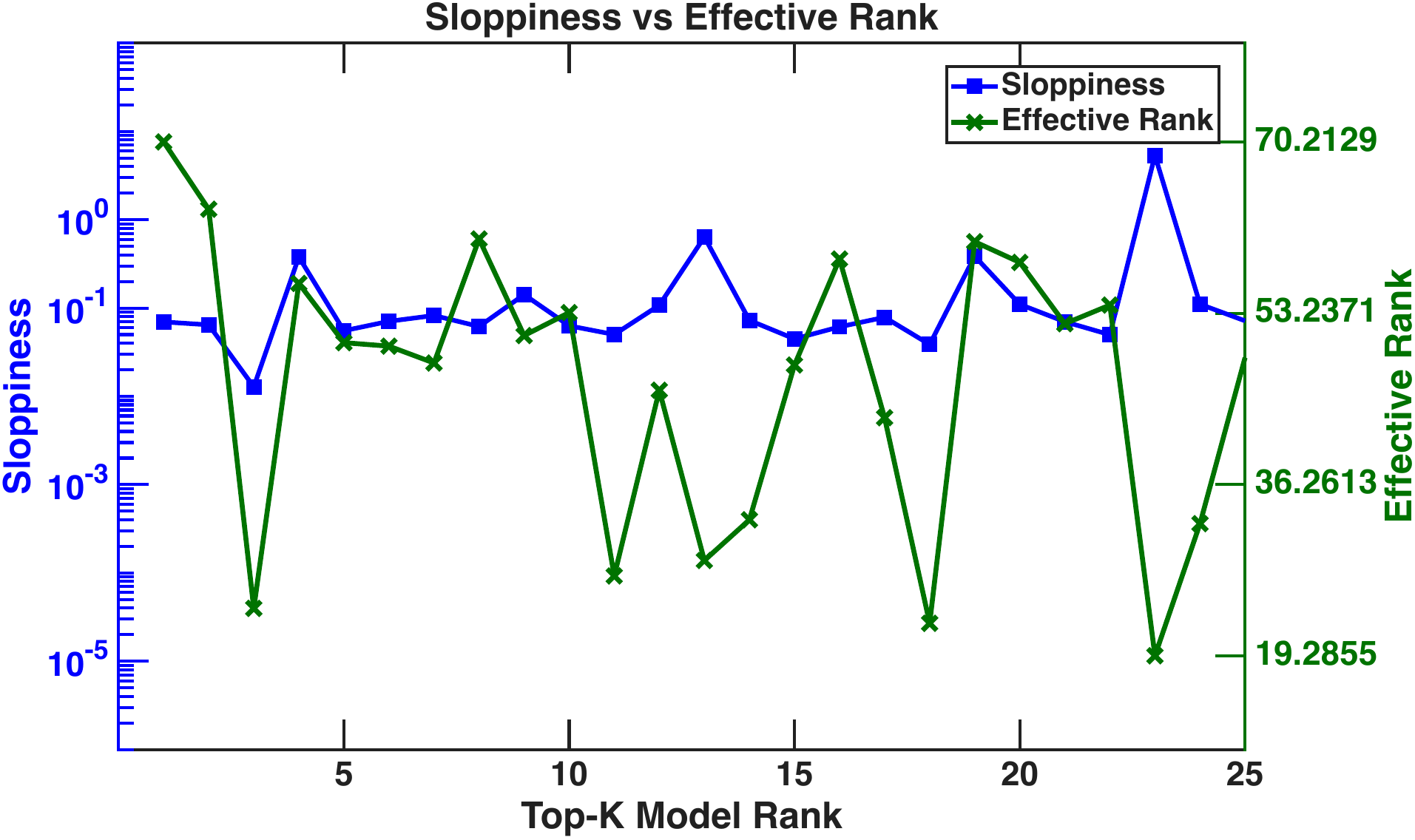}

\end{tabular}

\caption{
Columns correspond to the singular value ratio, effective rank, and Hessian eigenvalue spectrum, respectively. Rows illustrate representative models trained for the sine, exponential, and cubic target functions.
}

\label{fig:representative_results}

\end{figure}

\subsection{Reproducablity results}
This  summarizes the variability observed across multiple random initializations. The reported statistics correspond to the mean and standard deviation computed over independent training runs for the Top-$K$ functionally equivalent networks.

\begin{table*}[!ht]

\centering
\small

\caption{Functional non-uniqueness across different random initializations. Reported values correspond to the average minimum and average maximum observed among the Top-$K$ equivalent networks over multiple independent runs.}

\label{tab:initialization}

\renewcommand{\arraystretch}{1.3}

\resizebox{\textwidth}{!}{%

\begin{tabular}{llcccccc}

\toprule

Function &
Top-$K$ &
Neurons &
Parameters &
Effective Rank &
Sloppiness &
$R^2$ &
NRMSE \\

\midrule

\multirow{5}{*}{Sine}

&5
&45 -- 50
&137 --151
&$(1.8\pm0.1)\times10^{-5}$ --
$(2.0\pm0.2)\times10^{-4}$

&$(5.4\pm0.1)\times10^{-5}$ --
$(1.7\pm0.1)\times10^{-4}$

&$0.9998\pm10^{-6}$

&$0.004\pm10^{-5}$

\\

&10

&41 -- 50

&123 --151

&$(1.8\pm0.1)\times10^{-5}$ --
$(1.84\pm0.2)\times10^{-4}$

&$(5.4\pm0.1)\times10^{-5}$ --
$(1.7\pm0.1)\times10^{-4}$

&$0.9998\pm10^{-6}$

&$0.005\pm10^{-5}$

\\

&15

&5 --50

&16 --151

&$(1.8\pm0.1)\times10^{-5}$ --
$(2.6\pm0.3)\times10^{-3}$

&$(5.4\pm0.1)\times10^{-5}$ --
$(1.7\pm0.1)\times10^{-4}$

&$0.9997\pm10^{-5}$

&$0.006\pm10^{-4}$

\\

&20

&5 --50

&16 --151

&$(1.8\pm0.1)\times10^{-5}$ --
$(2.6\pm0.3)\times10^{-3}$

&$(5.4\pm0.1)\times10^{-5}$ --
$(1.7\pm0.1)\times10^{-4}$

&$0.9997\pm10^{-5}$

&$0.006\pm10^{-4}$

\\

&25

&5 --50

&16 --151

&$(1.8\pm0.1)\times10^{-5}$ --
$(2.6\pm0.3)\times10^{-3}$

&$(5.4\pm0.1)\times10^{-5}$ --
$(1.7\pm0.1)\times10^{-4}$

&$0.9997\pm10^{-5}$

&$0.006\pm10^{-4}$

\\

\bottomrule

\end{tabular}

}

\end{table*}

\FloatBarrier

\section{Additional Results for Multilayer Perceptrons}

This presents supplementary results obtained from the multilayer perceptron (MLP) experiments. The figures illustrate the geometric characteristics of the Top-$K$ functionally equivalent models under noisy training conditions together with numerical evidence supporting the proposed conjecture.

\subsection{Geometric Analysis under Noisy Conditions}

\begin{figure}[!ht]

\centering

\includegraphics[
width=\textwidth,
trim=0cm 16cm 0cm 16cm,
clip
]{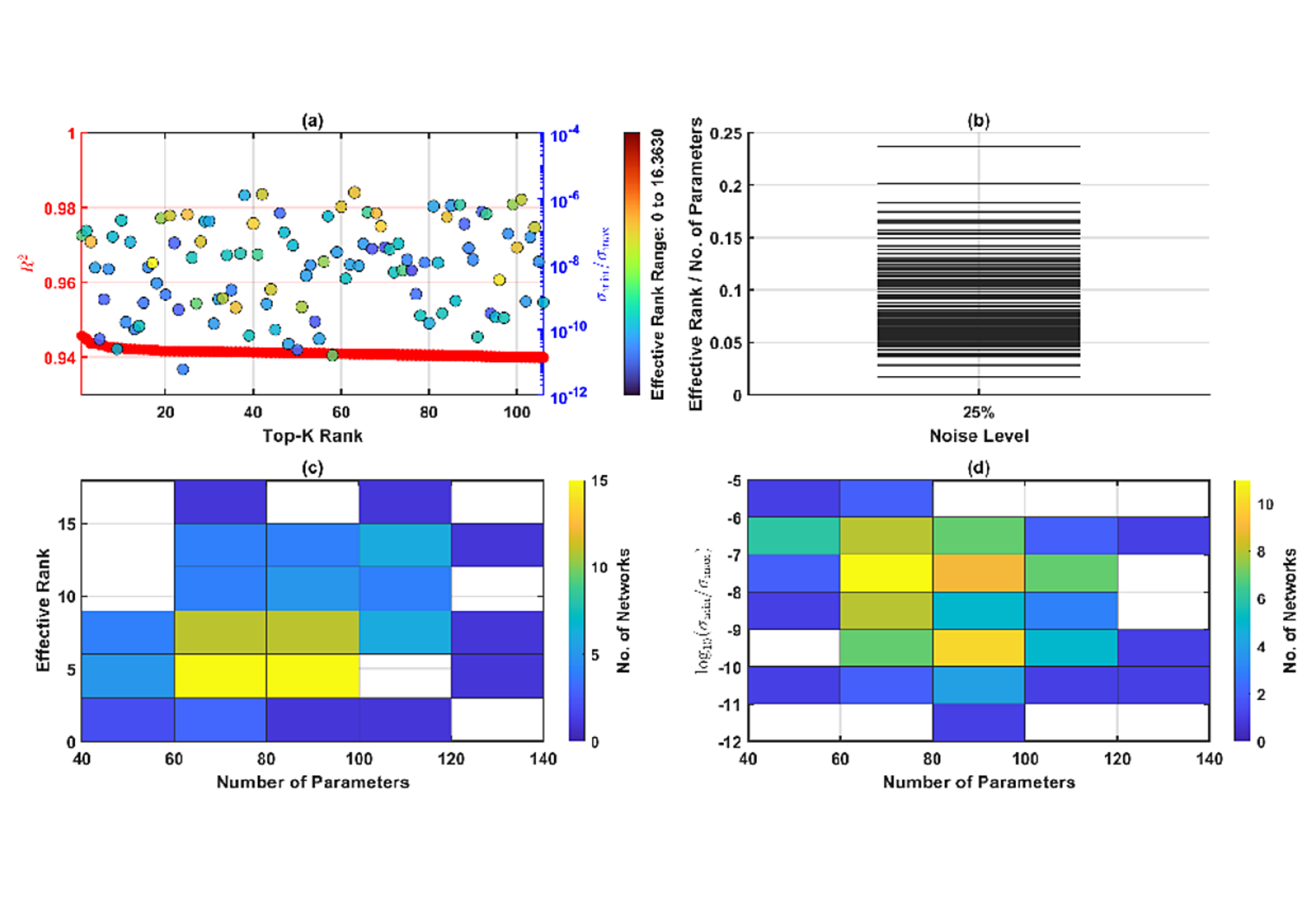}

\caption{Geometric analysis of the equivalence class obtained under noisy observations ($R^2\ge0.94$). (a) Distribution of equivalent networks ranked according to training $R^2$, together with the corresponding singular value ratio. (b) Ratio of effective rank to the total number of trainable parameters, illustrating the effective utilization of the parameter space. (c) Relationship between the number of trainable parameters and the effective rank. (d) Relationship between the number of trainable parameters and the logarithm of the singular value ratio.
}

\label{fig:noisy_geometry}

\end{figure}

To examine the effect of noise in functional equivalence, geometric diversity and model redundancy, the same set of 3905 MLP configurations are evaluated after adding gaussian noise to the data. In this case, the equivalence class consists of 106 networks with \(R^2 \geq 0.94\). Although the size of the equivalence class is smaller than in the noise-free case, there is still a significant number of networks that approximate the given noisy data.  Figure~\ref{fig:noisy_geometry} presents the geometric analysis of the equivalence class obtained under \(25\%\) noise. Figure~\ref{fig:noisy_geometry}(a) shows that for nearly identical \(R^2\) values, the networks have diverse geometric properties captured by effective rank and ratio of singular values. Figure~\ref{fig:noisy_geometry}(b) shows that the upper bound of the ratio of effective rank to number of parameters still remains $\approx$ 0.25, even while fitting noisy data. Figure~\ref{fig:noisy_geometry}(c) shows that the equivalent networks are mainly concentrated within the parameter range of 60 to 120 and effective rank ranging from 2.5 to 9. Figure~\ref{fig:noisy_geometry}(d) further shows that the equivalent networks are concentrated mainly within the parameter range of 40 to 120 with singular value ratio varying from \(10^{-11}\) to \(10^{-6}\) .

\FloatBarrier

\subsection{Numerical Evidence for the Proposed Conjecture}
\begin{figure}[!ht]
\centering

\includegraphics[
    width=\textwidth,
    trim=0cm 18cm 0cm 18cm,
    clip
]{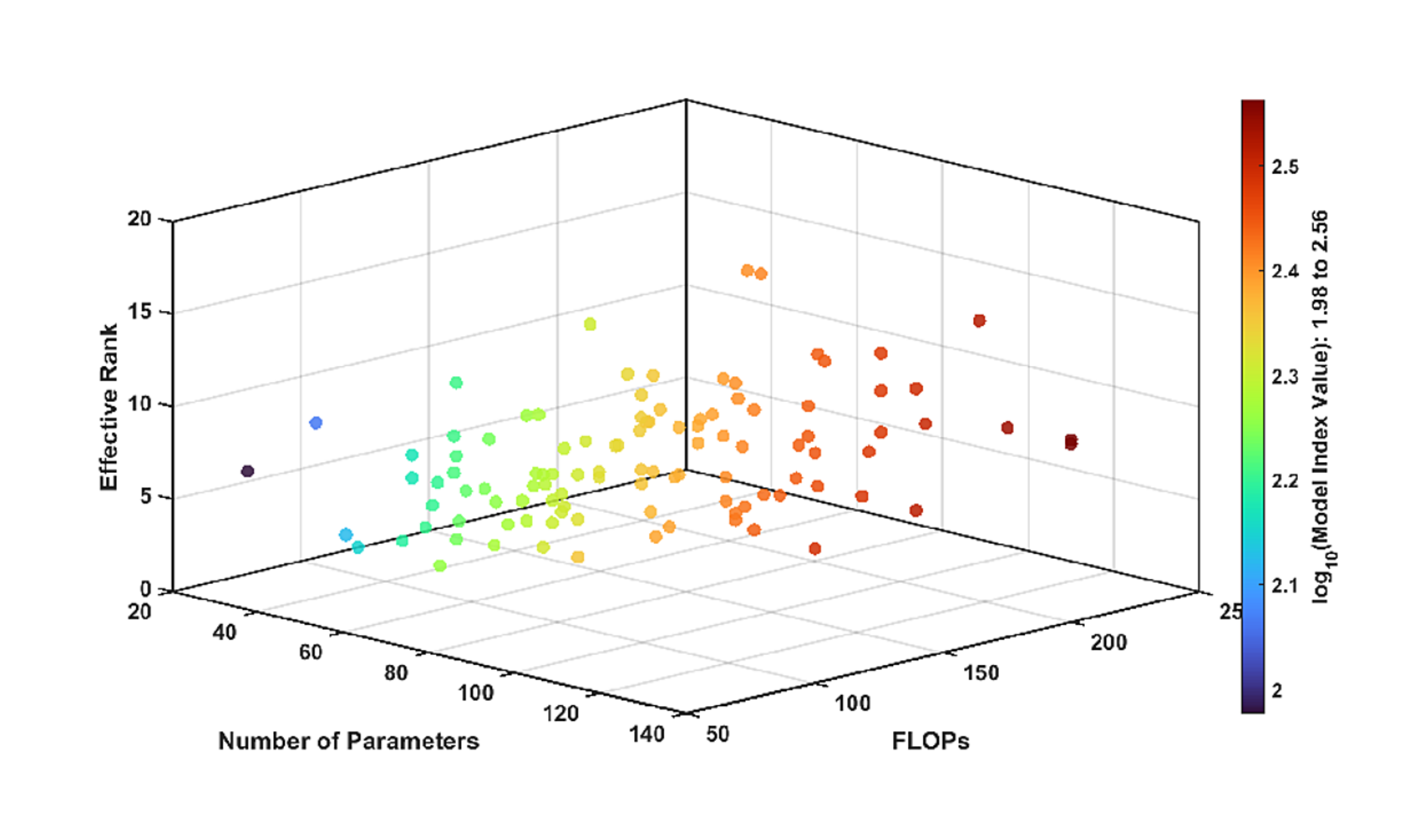}

\caption{
The three axes represent the number of trainable parameters, floating-point operations (FLOPs), and effective rank. The colour scale denotes the proposed model index, illustrating the trade-off between model complexity, computational cost, and effective parameter-space dimensionality across the equivalence class. Models with lower values of the proposed index are considered more parsimonious while maintaining comparable approximation performance.
}

\label{fig:conjecture_geometry}

\end{figure}

To numerically examine the proposed conjecture, we analyze models that belong to equivalence class in noise-free and noisy conditions. For each network, the number of trainable parameters and the corresponding FLOPs are computed, while the effective rank is estimated from the Hessian at optimal parameter vector. A network is said to be optimal in terms of parsimony, ease of estimation and inference energy when it minimizes the model index. The model index is computed for every architecture and plotted using a color gradient, where lower values (blue) indicate architectures that better satisfy the proposed criterion. As observed in Figure~\ref{fig:conjecture_geometry}, models that have low $\mathcal{M_I}$ are found to have low effective, less number of parameters and minimal FLOPs are indeed concentrated in the expected region.

\FloatBarrier

\section{Codes and saved results}

\subsection*{Source Code}

The complete MATLAB implementation, including scripts for data generation, neural network training, Hessian computation, effective rank estimation, and figure generation, is available at: \href{https://amritavishwavidyapeetham-my.sharepoint.com/:f:/g/personal/cb_ai_u4aim24005_cb_students_amrita_edu/IgCjtbvZPgj0R5xpw3HIfWkIATmL5BLz_bsVU6NAsI_B2Ys?e=zTZjoS}{Source Codes}

\subsection*{Experimental Results and Workspaces}

The complete collection of MATLAB workspaces, trained neural network models, intermediate experimental results, generated datasets, and supplementary analysis files required to reproduce the results presented in this paper is available at: \href{https://amritavishwavidyapeetham-my.sharepoint.com/:f:/g/personal/cb_ai_u4aim24005_cb_students_amrita_edu/IgAFvbhhARXTTJJl-5p-CWpwAXdXpuSWsrMZixrDGx7UdmY?e=hNWTki}{Experimental Results and Workspaces}

\end{document}